\documentclass[11pt]{article}

\usepackage[preprint]{acl}

\usepackage{times}
\usepackage{latexsym}
\usepackage{amsmath}

\usepackage[T1]{fontenc}
\usepackage[utf8]{inputenc}

\usepackage{microtype}

\usepackage{inconsolata}

\usepackage[greek,english]{babel}
\usepackage{textgreek}
\usepackage[LGR,T1]{fontenc}
\DeclareFontSubstitution{LGR}{cmr}{m}{n}
\usepackage{graphicx}
\usepackage{multirow}
\usepackage{booktabs}
\usepackage{longtable}   % allows multipage tables
\usepackage{tabularx}
\usepackage{float}
\usepackage[most]{tcolorbox}
\usepackage[dvipsnames]{xcolor}
\usepackage[table]{xcolor}
\usepackage{soul}

\usepackage{minted}
\setminted{
  frame=single,
  bgcolor=gray!10,
  fontsize=\footnotesize,
  autogobble,
  breaklines=true,
  breakanywhere=true,
  breaksymbolleft={},
  breaksymbolright={},
  breaksymbolindentleft=0pt,
  breaksymbolindentright=0pt,
  breaksymbolsepleft=0pt,
  breaksymbolsepright=0pt
}
\usepackage{algorithm}
\usepackage{algpseudocode}
\usepackage{xcolor}
\usepackage{listings}

\definecolor{backgray}{rgb}{0.97,0.97,0.97}
\definecolor{codegreen}{rgb}{0.20,0.45,0.25}
\definecolor{codepurple}{rgb}{0.50,0.20,0.55}
\definecolor{codegray}{rgb}{0.45,0.45,0.45}

\lstdefinestyle{compactpython}{
    language=Python,
    backgroundcolor=\color{backgray},
    basicstyle=\ttfamily\scriptsize,
    keywordstyle=\color{blue},
    commentstyle=\color{codegreen},
    stringstyle=\color{codepurple},
    numberstyle=\tiny\color{codegray},
    numbers=left,
    numbersep=4pt,
    frame=single,
    breaklines=true,
    breakatwhitespace=false,
    columns=fullflexible,
    keepspaces=true,
    showstringspaces=false,
    xleftmargin=0.5em,
    xrightmargin=0.5em,
    aboveskip=3pt,
    belowskip=3pt
}

\usepackage{tikz}
\usetikzlibrary{patterns.meta}

\definecolor{lightblue}{HTML}{D0D8F0} 
\DeclareRobustCommand{\hatchbox}[1]{%
  \tikz[baseline=(X.base)]{%
    \node[
      inner sep=2pt,
      outer sep=0pt,
      preaction={fill=white}, % white background
      pattern={
        Lines[
          angle=41,
          distance=8pt,      % larger hatch spacing
          line width=3pt
        ]
      },
      pattern color=lightblue
    ] (X) {\strut #1};
  }%
}

\newcommand{\framework}{\textsl{VISTA}} 

\newcommand{\best}[1]{\textbf{#1}}

\usepackage{wrapfig}

\title{\framework{}: A Versatile Interactive User Simulation Toolkit for Agent Evaluation} % AXIS — Agent eXploration through Interactive Simulation, HYDRA — HYbrid Dynamic Review for Agents

\author{
 \textbf{Yunan Lu\thanks{~~denotes equal contribution.}\textsuperscript{1,2}},
 \textbf{Ryan Shea$^{*}$\textsuperscript{1}},
 \textbf{Yusen Zhang\textsuperscript{1}},
 \textbf{Zhou Yu\textsuperscript{1,2}}
\\
 \textsuperscript{1}Department of Computer Science, Columbia University,
 \textsuperscript{2}Arklex.ai
\\
\texttt{\{yl4021, rs4235, yz5296, zy2461\}@columbia.edu}
}

\begin{document}
\maketitle
\begin{abstract}

Evaluation remains a critical bottleneck for interactive agent development. Existing evaluation methods often rely on static benchmarks, which fail to capture the dynamic, multi-step nature of agentic behavior and struggle to expose meaningful failure modes. While user-simulation-based evaluation offers a promising alternative, existing simulation frameworks suffer from two major limitations. First, they provide limited mechanisms for evaluating the quality and comprehensiveness of simulated interactions, making it difficult to assess whether a simulator sufficiently explores an agent’s capabilities and failure modes. Second, most frameworks are restricted to either UI-only actions or API-only actions, limiting their ability to model the full range of realistic user behaviors. To address these limitations, we propose \framework{}, a \textbf{V}ersatile \textbf{I}nteractive user \textbf{S}imulation \textbf{T}oolkit for \textbf{A}gent evaluation.
Our toolkit includes a suite of six metrics for measuring the realism, capability coverage, and interaction effectiveness of simulated interactions. In addition, we develop a hybrid user simulator that integrates both UI-based interactions and API-based interactions, enabling more realistic and comprehensive evaluation across diverse interactive environments. We evaluate \framework{} in e-commerce shopping and education customer service settings and demonstrate that it produces more realistic and comprehensive evaluations than existing methods.\footnote{Our code, documentation and demos are available at https://github.com/arklexai/arksim}

\end{abstract}

\section{Introduction}
\label{sec:introduction}

Interactive agents are increasingly deployed in real-world settings such as finance, education, and e-commerce \cite{dong-etal-2025-large, chu_llm_2026, AmazonAlexaShopping2026}. As these systems become more capable, evaluation has emerged as a critical development challenge. Existing evaluation approaches often rely on static benchmarks \cite{yehudai_survey_2026, mohammadi_evaluation_2025, liu_agentbench_2025, zhou_webarena_2024}, which measure performance on isolated tasks and overlook failure modes that arise during open-ended interactions such as poor action selection or tool-use  breakdowns. As a result, there remains a growing gap between benchmark performance and real-world agent reliability.

User-simulation-based evaluation has therefore emerged as a promising direction for scalable agent assessment. By using AI agents to emulate user behavior, simulators can generate multi-turn interactions without requiring extensive human participation \cite{dou_simulatorarena_2025, wang_human_2025, yao_-bench_2024, huang_crmarena_2024}. However, existing simulation frameworks provide limited mechanisms for evaluating the quality of the simulated interactions themselves, making it difficult to determine whether a simulator meaningfully exercises an agent’s capabilities. Prior work mainly focuses on assessing interaction fidelity \cite{zhu_realusersim_2026, zhou_mind_2026} and offers little insight into whether the generated interactions sufficiently explore the range of behaviors, capabilities, and failure modes relevant to the product agent.

In addition to limitations in evaluation methodology, existing simulation frameworks are constrained in the types of interactions they support. Most existing approaches support UI-based interactions \cite{lu_uxagent_2025, sun_llm_2025} or API-based actions \cite{barres_2-bench_2025}, but not both. UI-only simulators are often more error-prone and computationally expensive due to their need to handle noisy, long context information from raw web pages as opposed to short, structured information returned from API calls \cite{savadikar_shopgym_2026}. In contrast, simulators that rely solely on API-based actions are constrained by a limited set of tools and therefore unable to perform the full range of interactions needed to comprehensively evaluate modern agents. These limitations prevent existing approaches from fully capturing realistic user behavior and lead to incomplete evaluations. 
% of agent capabilities.

To address these limitations, we introduce \framework{}, a \textbf{V}ersatile \textbf{I}nteractive user \textbf{S}imulation \textbf{T}oolkit for \textbf{A}gent evaluation.
% a flexible evaluation toolkit for interactive agents based on hybrid user simulation.
Our toolkit introduces a suite of metrics for evaluating simulation quality, including measures of realism, capability coverage, and failure identification. These metrics provide insight into how comprehensively the simulator explores the agent’s functionalities and enable more rigorous and interpretable evaluation of interactive systems.

\begin{figure*}[!ht]
    \centering
    \includegraphics[scale=0.48]{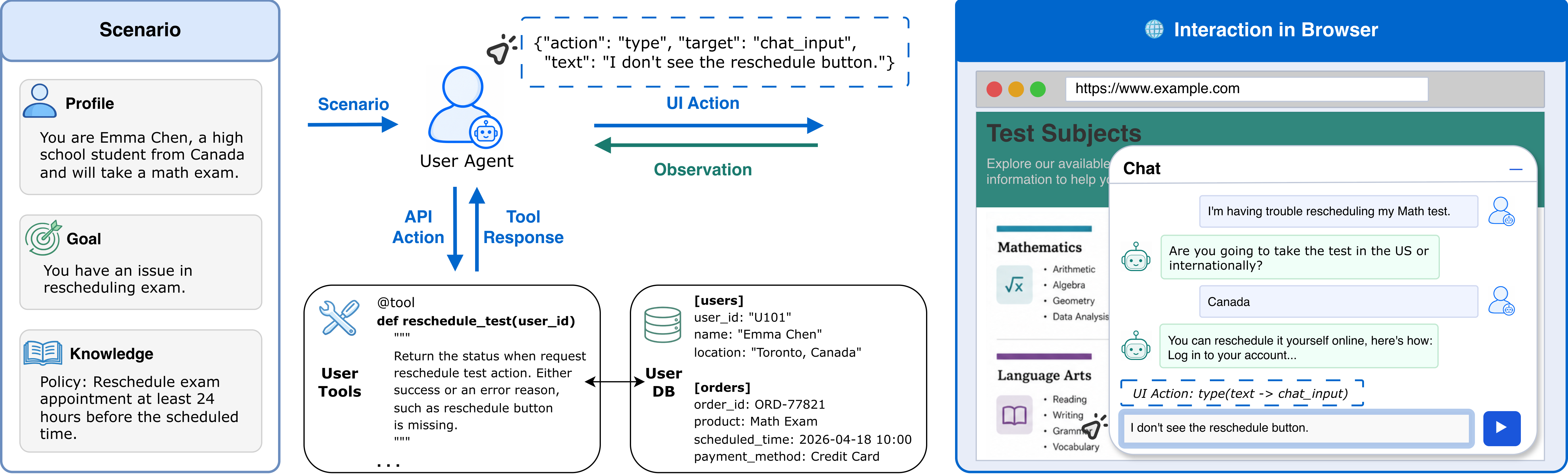}
    \caption{An overview of \framework{}. It configures each task scenario with a user profile, user goal, and relevant domain knowledge. During interaction, the user agent observes the webpage and decides whether to take UI actions directly or invoke user-side API tools for additional task-grounded information. The resulting interactions are then evaluated in terms of capability coverage, utterance realism, and interaction effectiveness, measured by agent failures.}
    \label{fig:pipeline}
    \vspace{-1.2em}
\end{figure*}

In addition, \framework{} introduces a hybrid user simulator capable of performing both UI-based and API-based operations alongside natural language interaction. Unlike prior approaches restricted to a single interaction paradigm, our hybrid framework enables more realistic simulation of user behavior across complex interactive environments. By combining conversational interaction with both UI-based and API-based actions, the simulator can evaluate a broader range of agent behaviors, including navigation, tool use, and action execution across multi-step workflows (see Figure \ref{fig:pipeline}).

% We apply \framework{} in two representative domains: e-commerce shopping and education. Our experiments demonstrate that the proposed toolkit produces more realistic and comprehensive evaluations than existing approaches. Furthermore, our proposed metrics provide meaningful signals regarding simulator quality and interaction coverage. Across both domains, \framework{} uncovers a broader range of agent failure modes and more effectively exercises agent capabilities, highlighting the importance of flexible hybrid simulation for interactive systems. Our contributions are as follows:

% \vspace{-0.4em}
% \begin{itemize}

% \item We introduce \framework{}, a flexible evaluation toolkit for interactive agents based on hybrid user simulation. Our toolkit includes a suite of six metrics for evaluating simulation quality, including measures of realism, capability coverage, and interaction effectiveness.
% \item We develop a hybrid user simulator that combines UI and API-based actions, enabling more realistic and comprehensive simulation of user behavior across environments.
% \item We empirically evaluate \framework{} in e-commerce shopping and education customer service settings, demonstrating that it produces more realistic and comprehensive evaluations while uncovering a broader range of agent failures than existing approaches.
% \end{itemize}

We apply \framework{} in two representative domains: e-commerce shopping and education. Our experiments demonstrate that the proposed toolkit produces more realistic and comprehensive evaluations than existing approaches. Furthermore, our proposed metrics provide meaningful signals regarding simulator quality and interaction coverage. Across both domains, \framework{} uncovers a broader range of agent failure modes and more effectively exercises agent capabilities. Importantly, \framework{} supports UI-only, API-only, and hybrid simulation, allowing practitioners to select the interaction mode that best matches their environment. Our contributions are as follows:

\vspace{-0.4em}
\begin{itemize}
% \vspace{-0.4em}
\item We introduce \framework{}, a flexible evaluation toolkit for interactive agents that supports UI-only, API-only, and hybrid user simulation. Our toolkit includes six metrics for evaluating simulation quality, including realism, capability coverage, and interaction effectiveness.
% \vspace{-0.4em}
\item We develop a hybrid user simulator that combines UI- and API-based actions, leveraging efficient API interactions when available while retaining the flexibility of UI interaction for broader user behavior coverage.
% \vspace{-0.4em}
\item We empirically evaluate \framework{} in e-commerce shopping and education customer service settings, demonstrating that it produces more realistic and comprehensive evaluations while uncovering a broader range of agent failures than existing approaches.
\end{itemize}

% \vspace{-0.4em}
\section{Related Work}
\label{sec:related}
% \vspace{-0.4em}
User simulation has emerged as a scalable alternative to human evaluation for interactive agents. Previous work has explored this paradigm across a variety of domains. For example, \citet{sun_llm_2025} investigate whether LLM agents can faithfully simulate customers interacting with shopping assistants, while \citet{lu_uxagent_2025} introduce UXAgent for automated usability testing of web interfaces. In business-oriented settings, \citet{shea_sage_2025} propose SAGE, a knowledge-grounded simulator that incorporates business logic to generate more realistic customer interactions. More recently, \citet{castelo_simgym_2026} demonstrate how user simulators can support scalable A/B testing and experimentation workflows.

Recent work has also explored how to evaluate the quality of the user simulators themselves. \citet{deng_mobile-bench_2024} introduce CheckPoint, a metric for measuring whether mobile agents reach key intermediate states during task execution, while \citet{zhou_mind_2026} propose the User-Sim Index to measure how closely simulated users resemble real human behavior. Although these metrics provide important insights, they are often tied to specific benchmarks or focus on human-likeness and task-state verification. Our work introduces general-purpose metrics that evaluate simulated interactions along dimensions such as realism, capability coverage, and interaction effectiveness, enabling a more holistic assessment of simulator quality.

Recent benchmarks have moved beyond purely conversational settings and begun evaluating agents in environments that require executable actions. Mobile-Bench augments mobile UI interactions with APIs to improve evaluation efficiency \citep{deng_mobile-bench_2024} and UXAgent enables browser-based interaction for usability testing \citep{lu_uxagent_2025}. $\tau^2$-Bench studies dual-control environments where both users and agents interact with a shared environment through tools \citep{barres_2-bench_2025}. These works demonstrate the importance of action-enabled evaluation, but existing approaches typically emphasize only UI-based actions or API-based actions. Our work builds on these ideas by introducing a hybrid simulation toolkit that combines both paradigms, allowing us to leverage the realism and flexibility of UI interaction alongside the efficiency and reliability of API-based actions.

\section{\framework{}}
\label{sec:our_method}

We introduce \framework{} as a flexible evaluation toolkit for interactive agents based on user simulation. Our toolkit includes a suite of evaluation metrics alongside a hybrid user simulator capable of taking both API-based actions and UI-based actions. We describe our evaluation metrics in Section \ref{sec:eval_metrics} and our hybrid user simulator in Section \ref{sec:simulator_method}.

\subsection{Evaluation Metrics}
\label{sec:eval_metrics}

\framework{} includes six metrics spanning coverage, realism, cost, and failure identification. These metrics are described in detail in the following sections.

% \subsubsection{Assistant Agent}
% Beyond evaluating the user agent directly, we examine the behavior it elicits in the assistant agent. A strong user simulator should expose the assistant to varied tool calls and heterogeneous tool-call sequences, which are more likely to trigger unique failure cases. We therefore evaluate assistant-agent outcomes from two perspectives: coverage, which measures how broadly the simulator exercises the assistant's tool-use behavior, and error identification, which measures how effectively it reveals distinct assistant-agent failures. We evaluate coverage using with tool-calling transition entropy and tool-calling frequency entropy, while error identification is measured using a LLM-as-a-Judge approach. Each metric is described in detail below.

\subsubsection{Coverage}

Coverage measures whether the user agent can drive the product agent through diverse tool-use patterns, rather than repeatedly triggering similar execution paths. We evaluate coverage using three complementary metrics: tool-calling transition entropy, tool-calling distribution entropy, and trajectory distance.

\paragraph{Tool-calling Transition Entropy}
To measure the diversity of the product agent's tool-calling sequences, we propose \textit{TransitionEntropy} (TE), which measures the normalized entropy of the tool-calling transition matrix. For each interaction trajectory, we extract adjacent tool-call pairs and aggregate them into an $N \times N$ transition matrix, where $N$ is the number of agent tools. Each row defines the empirical distribution of the next tool conditioned on the current tool. We compute the Shannon entropy of each non-empty row, normalize it by $\log_2 N$, and average across valid rows:
% \[
% \mathrm{TransitionEntropy}
% =
% \frac{1}{|\mathcal{R}|}
% \sum_{i \in \mathcal{R}}
% \frac{
% -\sum_{j=1}^{N} p_{ij}\log_2 p_{ij}
% }{
% \log_2 N
% }
% \]
\[
\resizebox{0.95\columnwidth}{!}{$
\displaystyle
\mathrm{TransitionEntropy}
=
\frac{1}{|\mathcal{R}|}
\sum_{i \in \mathcal{R}}
\frac{
-\sum_{j=1}^{N} p_{ij}\log_2 p_{ij}
}{
\log_2 N
}
$}
\]
where $p_{ij}$ is the probability of calling tool $j$ after tool $i$, and $\mathcal{R}$ is the set of tools with at least one outgoing transition. The score ranges from $0$ to $1$. A low score indicates that the conversation follows a rigid, predictable conversation flow with less diversity, while a high score indicates flexible routing across different next-tool choices.

\paragraph{Tool-calling Distribution Entropy}

Transition entropy measures local routing diversity, but does not fully capture global tool coverage. Therefore, we additionally measure the tool-calling distribution entropy, \textit{ToolDistrEntropy} (TDE), where $q_i$ denotes the empirical frequency of tool $i$ among all agents' tool calls. We compute the normalized Shannon entropy:
\[
\resizebox{0.75\columnwidth}{!}{$
\displaystyle
\mathrm{ToolDistrEntropy}
=
\frac{
-\sum_{i=1}^{N} q_i \log_2 q_i
}{
\log_2 N
}
$}
\]
This score also ranges from $0$ to $1$. A score near $0$ means that the agent relies on only one or a few tools, while a score near $1$ means that tool calls are distributed evenly across the available tools. This metric complements transition entropy by capturing global tool coverage rather than sequential routing diversity.

\paragraph{Trajectory Distance}
Both transition entropy and distribution entropy measure aggregate tool-calling diversity on tool names across the full dataset. High scores indicate broad tool coverage and diverse tool transitions, but they do not rule out repeated trajectories under the same user goal. Such repetition reduces both diversity and efficiency. We propose \textit{TrajDist} (TD) to measure within-goal trajectory diversity over tool names and tool arguments.

For each interaction trajectory, we represent the agent's tool-use behavior as a sequence of $(\texttt{tool\_name}, \texttt{input\_repr})$ pairs, where $\texttt{input\_repr}$ is a serialized representation of the tool arguments. Within each user goal group, we compute the average pairwise normalized Levenshtein distance between all trajectory pairs:
\[
\resizebox{0.67\columnwidth}{!}{$
\displaystyle
\mathrm{TrajDist}
=
\frac{1}{N}
\sum_{a < b}
\frac{
\mathrm{Lev}(s_a, s_b)
}{
\max(|s_a|, |s_b|)
}
$}
\]
where $N$ is the number of pairs in the group, which is $\binom{|\mathcal{T}|}{2}$, $s_a$ and $s_b$ are two tool-call sequences, and $\mathrm{Lev}(\cdot,\cdot)$ denotes the Levenshtein distance between two sequences.

The score ranges from $0$ to $1$. A score near $0$ means that trajectories within the same user goal group follow nearly identical tool-call paths with similar arguments. A score near $1$ means that the trajectories differ substantially in either tool choices, order, or arguments. 
% This metric complements the two entropy-based metrics by capturing fine-grained behavioral variation beyond tool-type coverage.

\subsubsection{Realism}
% Real human-agent interactions are often inaccessible for deployed product agents due to privacy and availability constraints. Therefore,
We evaluate the user simulator's realism with reference-free LLM-as-a-Judge scores. Following prior work, we assess whether each generated user utterance is consistent with three behavior components from its prompt: the user profile, task goal, and domain knowledge \cite{shea_sage_2025}. The details of these components are in Section \ref{sec:gen_scenario}. Since the components are either manually derived from real interactions between customer-agent conversations or manually configured by domain experts, they provide realistic behavioral constraints. 
% However the simulator may not always follow them. 
% To fill in this gap, we use an LLM-as-a-judge to determine whether each utterance is consistent with the assigned profile, goal, and knowledge,
Therefore we report the corresponding consistency rates as automatic proxies for behavior realism. The judging prompt is provided in Appendix \ref{sec:app_prompt_realism}.

\subsubsection{Cost}
We evaluate the efficiency of user simulator using two cost metrics. First, we use the average number of prompt and completion tokens per turn, which serves as a proxy for monetary cost. Second, we use the average number of actions per turn as a proxy for time cost. We use this proxy instead of direct elapsed time because wall-clock measurements can vary with implementation details, such as parallel execution and system load. Actions include both API calls and UI operations, each of which contributes to the time required to generate an interaction. Therefore, a simulator with more actions per turn generally incurs higher time cost.

\subsubsection{Agent Failure Identification}
We also assess the user agent's ability to identify agent failures. 
Following the prior work \cite{shea_sage_2025}, we use an LLM-as-a-Judge to evaluate each assistant response along five dimensions: helpfulness, coherence, verbosity, relevance, and faithfulness. For each dimension, the judge returns a score from 0 to 4 with a brief justification. 
We pass the full conversation and the scores to a second LLM, which consolidates the justifications into a set of unique agent failures. We use the number of unique failures as the failure-identification metric. 
A higher count indicates broader ability to evaluate the agent’s capabilities and failure cases.

\subsection{Hybrid User Simulation}
\label{sec:simulator_method}
\framework{} incorporates a hybrid user simulator that can take both API-based and UI-based actions. This allows the simulator to improve interaction quality by combining the increased flexibility of UI-based simulation with the enhanced reliability and cost effectiveness of API-based simulation. \framework{} begins by constructing task scenarios, which define the context and objective of each simulated interaction (Section~\ref{sec:gen_scenario}). It then equips the simulator with API-based and UI-based actions (Sections~\ref{sec:user_tools} and~\ref{sec:predict_UI}).
Finally, it integrates API-based tool responses into UI action prediction through an iterative observation-planning-action loop (Section~\ref{sec:orchestrate_hybrid}).
An overview of the full process is provided in Figure~\ref{fig:pipeline}.

\subsubsection{Generate Scenarios}
\label{sec:gen_scenario}
To produce realistic and diverse interactions, we construct each simulation scenario from three components: a user profile, a user goal, and relevant domain knowledge, following prior work \cite{shea_sage_2025}. The user profile captures attributes that influence user behavior, including demographics, preferences, personality traits, and behavioral patterns. The user goal defines the task the simulated user is trying to complete, such as finding a product or requesting help with a course concept. We design these goals to cover the capabilities of the product agent. The domain knowledge defines the information context available to the user, such as webpages, FAQs, product catalogs, or other publicly available resources. 
We combine these three components in a structured template to construct the system prompt, which guides the user simulator throughout its interaction with the product agent. See Figure \ref{fig:pipeline} for an example.

\subsubsection{Create User API Tools}
\label{sec:user_tools}
To make the user simulator reusable across domains while remaining closely aligned with specific product agents, we create two categories of user API tools: general tools and task-specific tools. 
% Our test cases are specifically designed for interactive, customer-facing agents however our toolkit can incorporate a variety of tools as specified by the user.
Our test cases are specifically designed for interactive, customer-facing agents however our toolkit can incorporate a variety of tools to simulate domain-specific user behavior as specified by the user.

General tools are shared across domains and scenarios. These include web search, web crawling, and retrieval from knowledge bases or user preference memory. Such tools allow the simulator to behave like a real user who can look up information, recall preferences, and inspect external resources to answer clarification questions or proactively explore information for follow-up questions.

Task-specific tools are customized for the product agent being evaluated. 
% For an e-commerce shopping assistant, such tools include \texttt{get\_order\_info}, which allows the simulator to answer dynamic questions such as “What is your order ID?”. Additional, user tools include not only read but also write operations. 
% For a technical support agent, users often follow the agent’s instructions and then report the outcome. 
% To model this behavior, the user simulator needs tools that allow it to execute the corresponding action and return the resulting status, error message, or data to the agent. 
For an e-commerce shopping agent, these tools include things such as retrieving order information or canceling/returning orders. 
For an education agent, the tools may include actions such as login, rescheduling a test, or launching an exam.

To scalably create diverse task-specific tools, we implement them as mocked APIs with varied responses. Each response includes structured status fields, returned data, and realistic error messages. We derive the requested data from real database schema in JSON format and align error messages with those shown in the user interface. Mocked APIs decouple the simulator from production systems while preserving realistic interaction dynamics. This design enables large-scale testing without modifying production databases, affecting live user interfaces, or manually configuring diverse test accounts. For example, an exam rescheduling task may need to handle cases where the request occurs within one hour of the exam, the new location is in an unsupported country, or the selected test type is not eligible for rescheduling. 
% Mocked user tools allow us to generate diverse user states and execution outcomes programmatically, enabling evaluation across a broad range of realistic conditions.

\subsubsection{Predict UI Actions}
\label{sec:predict_UI}
To interact with the product agent through its deployed interface, we adopt UXAgent's Universal Web Connector \cite{lu_uxagent_2025, sun_llm_2025} as action module for UI action prediction.
Given the DOM tree observation, previous interactions and additional context from API tools, the connector prompts an LLM to generate the next executable action in JSON format. The output includes the action type, the target element ID, and any additional fields required by the specific action type. The supported action types include \texttt{click}, \texttt{type}, \texttt{select}, \texttt{key\_press}, \texttt{clear}, \texttt{scroll}, \texttt{navigation}, and \texttt{terminate}. 
For example, if the user simulator decides to send a message to the chatbot, it may output: \{\texttt{"action"}: \texttt{"type"}, \texttt{"target"}: \texttt{"chat\_input"}, \texttt{"text"}: \texttt{"I don't see the reschedule button."}\}.
The predicted action is then executed on the webpage, producing the next observation for the following interaction step.

\subsubsection{Orchestrate Hybrid Actions}
\label{sec:orchestrate_hybrid}
The hybrid user simulator interacts with the product agent through an iterative loop of three modules: observation, planning, and action.
At each iteration, the observation module captures the DOM tree through UXAgent.
Given the user's goal and provided user tools (Section~\ref{sec:user_tools}), the planning module uses an LLM-based router to decide whether to call an API tool or perform a UI-action (see Appendix \ref{sec:app_prompt_user_agent} for the prompt).
If choose API tool, it further predicts the tool name and its arguments. 
Finally, the executed tool response is incorporated into the UI action predictor, which produces the next UI action as described in Section~\ref{sec:predict_UI}. 

The interaction continues through this observation-planning-action loop until the user agent outputs a \texttt{terminate} action or reaches the predefined turn budget. After the interaction ends, we extract the conversation history from the chat widget and use it for downstream evaluation of the product agent’s performance.

% \vspace{-0.4em}
\section{Experiments}
\label{sec:experiments}
% \vspace{-0.4em}

We use the \framework{} toolkit to evaluate two customer-facing agents: an e-commerce shopping agent and an education agent. In both domains, we evaluate three user-simulation configurations: \textbf{API-only}, \textbf{UI-only}, and \textbf{hybrid}. The API-only simulator uses the available user-side API tools to retrieve information and execute task-specific operations, but does not navigate the user-facing interface. The UI-only simulator performs user-side information retrieval and task execution exclusively through the deployed web interface. Our hybrid simulator integrates both types of actions. 
All three simulators communicate with customer-facing agent through natural language.
% \vspace{-0.4em}
\paragraph{E-commerce Shopping Agent.}
We simulate interactions with a shopping agent deployed on the Shopify platform. The agent supports both product exploration and post-purchase services through 13 APIs, including 8 read operations and 5 write operations. Representative tools include \texttt{get\_recommendation}, \texttt{get\_customer\_info}, \texttt{get\_product}, \texttt{return\_item}, and \texttt{cancel\_order}. The agent also has access to FAQs and store policies, which support policy-grounded responses. For this use case, the hybrid user simulator has access to general user API tools (Section \ref{sec:user_tools}) and a task-specific \texttt{get\_customer\_info} tool.
% For the hybrid user simulator design in this usecase, besides the general user tools, i.e., web search, web crawling, and retrieval, it includes the task-specific tool \texttt{get\_customer\_info}. It allows the simulator to retrieve customer- and order-related context and generate grounded utterances efficiently.
% \vspace{-0.4em}
\paragraph{Education Agent.}
We also generate interactions with a customer-service agent deployed on an international education platform. The agent provides technical support for student-facing issues, such as login problems, exam rescheduling, test launch failures, and questions about exam policies, diplomas, and region-specific requirements. To support diverse user operations across troubleshooting scenarios, we create eight mocked user tools for the hybrid user simulator, such as \texttt{login}, \texttt{edit\_account}, and \texttt{cancel\_test}. See Appendix~\ref{sec:app_edu_tools} for details.

To demonstrate the utility of \framework{}'s hybrid simulator and the usefulness of its evaluation metrics, we conduct experiments with LLMs from both commercial and open-source model families: \texttt{gpt-5.4}~\cite{singh2026openaigpt5card} and \texttt{Qwen3.5-27B}~\cite{qwenteam2026qwen35omnitechnicalreport}. 
For each model and use-case, we generate 100 interactions under the API-only, UI-only, and hybrid conditions, resulting in 300 interactions.
Using these interactions, we perform  evaluations based on the metrics described in Section \ref{sec:eval_metrics}. This setup allows us to examine whether the hybrid simulator can generate higher-quality interactions than existing baselines and if our proposed metrics can capture performance differences across simulator backbones.

% \vspace{-0.4em}
\subsection{Human Evaluation}
% \vspace{-0.4em}
Additionally, we conduct a human evaluation to assess the realism of all three simulator configurations and to provide a point of comparison for our proposed automatic metrics. We recruit two annotators to evaluate 300 simulated interactions in total: 150 interactions with the education agent and 150 interactions with the shopping agent. For each domain, we sample 50 interactions from each of the API-only, UI-only, and hybrid conditions. Annotators rate the simulated-user utterances on a 3-point Likert scale along three dimensions:

\begin{itemize}
\vspace{-0.5em}
\item \textbf{Human-likeness}: How closely the simulated user’s language resembles that of a human.
\vspace{-0.5em}
\item \textbf{Coherence}: How well the simulated user’s utterances follow logically from the history.
\vspace{-0.5em}
\item \textbf{Consistency}: How consistent the simulated user is with its goal during the conversation.
\vspace{-0.5em}
\end{itemize}

We conduct a separate human evaluation specifically for the education agent. We recruit four annotators to evaluate 300 interactions: 75 interactions from each of the API-only, UI-only, and hybrid simulators, together with 75 interactions between the education agent
and real customers. For each conversation we have the annotators rate whether they felt the user utterances came from a human, a bot, or if they weren't sure. All conversations for our human evaluations generated with \texttt{gpt-5.4}. See Appendix \ref{sec:app_human_eval} for details.

\begin{table*}[t]
\centering

% \scriptsize
\fontsize{8.5}{9.5}\selectfont
\setlength{\tabcolsep}{3.2pt}
\renewcommand{\arraystretch}{0.95}

\begin{tabular*}{\textwidth}{
  @{\extracolsep{\fill}}
  llccccccccc
  @{}
}
\toprule
\multirow{2}{*}{\textbf{Model}} &
\multirow{2}{*}{\textbf{Method}} &
\multicolumn{3}{c}{\textbf{Coverage} $\uparrow$} &
\multirow{2}{*}{\textbf{\# Failures}} &
\multicolumn{3}{c}{\textbf{Realism (\%)} $\uparrow$} &
\multicolumn{2}{c}{\textbf{Cost/turn} $\downarrow$} \\
\cmidrule(lr){3-5}
\cmidrule(lr){7-9}
\cmidrule(l){10-11}
& & \textbf{TE} & \textbf{TDE} & \textbf{TD}
& & \textbf{Goal} & \textbf{Profile} & \textbf{Knowledge}
& \textbf{Tokens} & \textbf{Actions} \\
\midrule

\multicolumn{11}{@{}l}{\textbf{E-commerce Shopping Agent}} \\
\midrule
\texttt{gpt-5.4}
& API-only & 0.33 & 0.57 & 0.93 & 64 & 98.95 & 96.52 & 94.43 & \best{0.5k} & \best{1.05} \\
& UI-only  & 0.33 & 0.59 & 0.94 & 69 & 99.32 & 97.97 & 92.20 & 79.9k & 3.18 \\
& Hybrid   & \best{0.34} & \best{0.62} & \best{0.96} & \best{79}
& \best{99.40} & \best{99.39} & \best{94.58} & 69.1k & 2.03 \\
\midrule

\texttt{Qwen3.5-27B}
& API-only & 0.26 & 0.54 & 0.92 & 48 & 95.50 & 92.59 & 94.71 & \best{0.4k} & \best{1.03} \\
& UI-only  & 0.27 & 0.55 & 0.94 & 51 & 96.39 & 93.98 & 96.39 & 57.1k & 2.59 \\
& Hybrid   & \best{0.29} & \best{0.57} & \best{0.97} & \best{55}
& \best{96.74} & \best{94.46} & \best{96.42} & 56.6k & 2.33 \\
\midrule

\multicolumn{11}{@{}l}{\textbf{Education Agent}} \\
\midrule
\texttt{gpt-5.4}
& API-only & 0.16 & 0.80 & 0.63 & 30 & 97.83 & 97.10 & 95.29 & \best{0.6k} & \best{1.18} \\
& UI-only  & 0.16 & 0.78 & 0.62 & 24 & 99.72 & 96.97 & 94.21 & 66.7k & 1.76 \\
& Hybrid   & \best{0.17} & \best{0.81} & \best{0.68} & \best{34}
& \best{100.00} & \best{97.46} & \best{96.06} & 86.0k & 1.73 \\
\midrule

\texttt{Qwen3.5-27B}
& API-only & \best{0.11} & 0.77 & 0.67 & 26
& 97.25 & 96.14 & 96.14 & \best{0.7k} & \best{1.25} \\
& UI-only  & \best{0.11} & 0.77 & 0.64 & 23
& 98.35 & 94.24 & \best{96.71} & 64.6k & 2.74 \\
& Hybrid   & \best{0.11} & \best{0.78} & \best{0.68} & \best{30}
& \best{98.68} & \best{98.24} & 96.48 & 86.6k & 2.82 \\
\bottomrule
\end{tabular*}
\caption{Comparison of API-only, UI-only, and hybrid user simulators
built on commercial and open-source models. Coverage includes transition
entropy (TE), tool-distribution entropy (TDE), and trajectory distance
(TD). Realism measures consistency with the user goal, profile, and domain
knowledge. Cost is reported per turn. Best results are shown in
\textbf{bold}.}
\label{tab:main}
\end{table*}

\begin{figure*}[t]
\centering
% ---------- Left: Table ----------
\begin{minipage}[t]{0.49\textwidth}
\vspace{0pt}
\centering

\small
\setlength{\tabcolsep}{3.5pt}
\renewcommand{\arraystretch}{0.95}
\begin{tabular*}{\linewidth}{
    @{\extracolsep{\fill}}lccc@{}
}
\toprule
\textbf{Method}
& \shortstack{\textbf{Human-}\\\textbf{likeness}}
& \textbf{Coherence}
& \shortstack{\textbf{Goal}\\\textbf{Consistency}} \\
\midrule
\multicolumn{4}{@{}l}{\textbf{E-commerce Shopping Agent}} \\
\midrule
API-only
& 1.92 (0.34)
& 2.76 (0.51)
& 2.60 (0.44) \\
UI-only
& 1.97 (0.56)
& 2.79 (0.41)
& 2.60 (0.59) \\
Hybrid
& \best{2.04 (0.49)}
& \best{2.80 (0.40)}
& \best{2.63 (0.61)} \\
\midrule
\multicolumn{4}{@{}l}{\textbf{Education Agent}} \\
\midrule
API-only
& 2.10 (0.58)
& 2.68 (0.47)
& 2.36 (0.48) \\
UI-only
& 2.18 (0.48)
& 2.70 (0.54)
& 2.16 (0.79) \\
Hybrid
& \best{2.26 (0.66)}
& \best{2.74 (0.52)}
& \best{2.38 (0.69)} \\
\bottomrule
\end{tabular*}
\captionof{table}{Human evaluation on a 3-point Likert scale. Better scores are shown in
\textbf{bold}; standard deviations are in parentheses. \\}
\label{tab:human_eval}
\end{minipage}
\hfill
% ---------- Right: Figure ----------
\begin{minipage}[t]{0.46\textwidth}
\vspace{0pt}
\centering
\includegraphics[
    width=0.90\linewidth
]{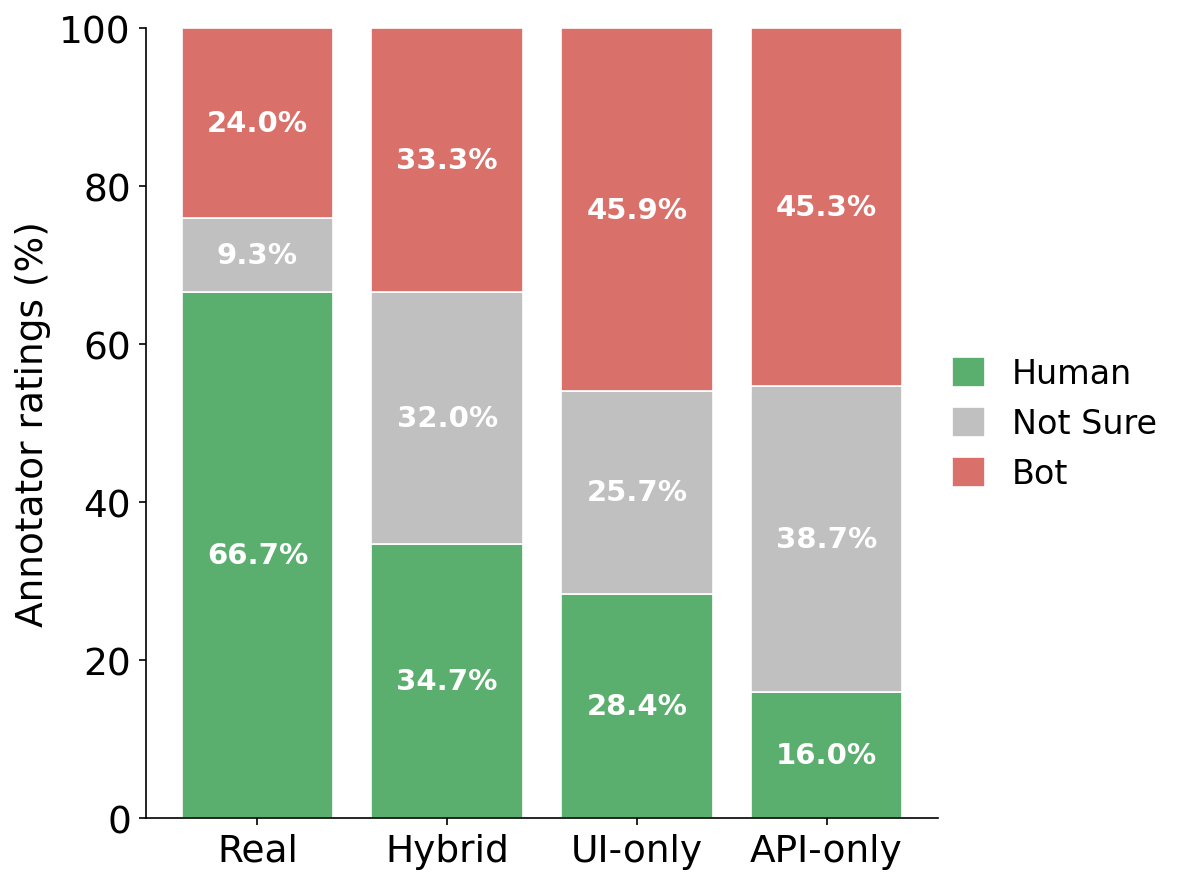}
\vspace{-0.6em}
\captionof{figure}{Annotator classifications of user messages from interactions between the education agent and real customers, API-only, UI-only, and hybrid simulators.}
\label{fig:human_eval_ged}
\end{minipage}
% \vspace{-1em}
\end{figure*}
% \vspace{-0.4em}
\section{Results}
\label{sec:results}
% \vspace{-0.4em}
The results of our experiments are given in the following sections. We find that our hybrid user simulator consistently outperforms the API-only and UI-only simulator according to both our automatic metrics and human evaluations. We also see that our proposed metrics are able to capture meaningful differences between simulation methods, model type, and correlate with human judgments.

\subsection{Hybrid Simulator Performance}

% \vspace{-0.4em}
\paragraph{Automatic Evaluation}
The results of our automatic evaluation are in Table \ref{tab:main}. Across both domains and model backbones, the hybrid simulator consistently outperforms the API-only and UI-only simulators on coverage metrics by up to 12\% and identifies up to 42\% more unique agent failures. 
These improvements suggest that combining UI-based and API-based actions enables the simulator to explore a broader range of agent capabilities and interaction trajectories. The hybrid simulator also achieves high realism scores, including 100\% goal consistency in the education agent case, while maintaining comparable operational cost to the UI-only setup. Although the hybrid framework occasionally incurs higher token usage due to the additional prompting required to route between UI and API interaction modes, the action cost remains similar to the UI-only setup. The API-only simulator is substantially less expensive because it operates on short, structured tool responses without processing webpage observations. However, it cannot reproduce interaction patterns requiring UI navigation. The hybrid simulator therefore uses efficient API interactions when available while retaining UI flexibility, enabling more comprehensive agent evaluation.

% \vspace{-0.4em}
\paragraph{Human Evaluation}
These improvements are further supported by the human evaluation results shown in Table \ref{tab:human_eval}. Across both domains, the hybrid simulator receives consistently higher ratings than both the API-only and UI-only simulators for human-likeness, coherence, and goal consistency while also being 6.3\% more likely to be mistaken for a human than the UI-only baseline and 18.7\% more likely than the API-only baseline (Figure \ref{fig:human_eval_ged}). We believe these improvements stem from the hybrid simulator’s ability to more reliably retrieve and utilize structured user information during interactions while retaining the flexibility of UI interaction. While the UI-only simulator has access to the same information through interface interactions, retrieving it through UI actions is often less reliable and more prone to navigation or execution errors. In contrast, the API-only simulator can reliably retrieve structured information but cannot get up-to-date information from web observation. As a result, both single-modality baselines may produce less realistic responses: the UI-only simulator more frequently falls back to generic or artificial responses, whereas the API-only simulator cannot ground its behavior in the state of the user interface.

\begin{figure*}[t]
    \centering
    \includegraphics[scale=0.48]{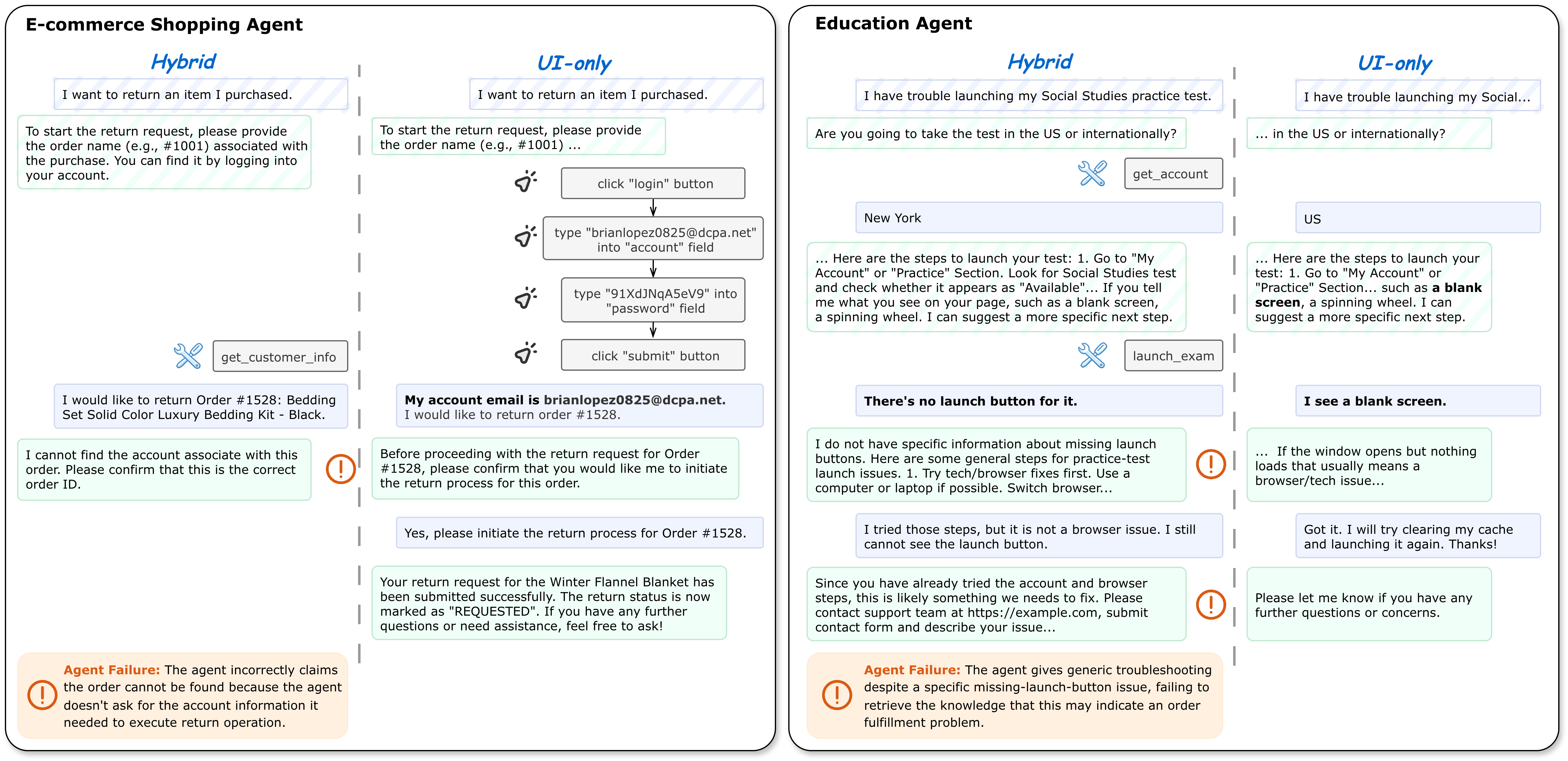}
    \caption{Interaction samples generated by the UI-only and hybrid user simulator under the same task scenarios: returning an item for the e-commerce shopping agent and troubleshooting an exam launch issue for the education agent. {\setlength{\fboxsep}{1pt}\colorbox[HTML]{D0D8F0}{Blue messages}} are from user simulator, and {\setlength{\fboxsep}{1pt}\colorbox[HTML]{C0E8D0}{green messages}} are from the product agent. \protect\hatchbox{Hatched messages} indicate content is identical across both settings. \textbf{Bold} text highlights the key divergence between the two simulators that determines whether an agent failure is triggered. See Appendix \ref{sec:app_agent_errors} for more agent failures.}
    \label{fig:case_study}
    \vspace{-1.2em}
\end{figure*}

% \vspace{-0.4em}
\subsection{Simulation Evaluation Metric Performance}
% \vspace{-0.4em}
Our experiments demonstrate that the proposed evaluation metrics provide meaningful signals about simulator quality. As shown in Table \ref{tab:main}, the larger \texttt{gpt-5.4} model generally achieves stronger performance than the smaller model across both domains, suggesting that stronger language models better maintain goal consistency, profile consistency, and knowledge grounding during simulation.

Our proposed realism metrics also capture descriptive differences among simulation methods. For the \texttt{gpt-5.4} simulator, the hybrid setup generally achieves higher scores across most realism dimensions. This pattern is also observed in Table \ref{tab:human_eval}, where the hybrid simulator receives the highest average human ratings across all three dimensions.

Our proposed coverage metrics also show strong agreement with downstream measures of evaluation effectiveness. Higher coverage scores are consistently associated with more unique agent failures identified during evaluation. The hybrid simulator achieves equal or higher transition entropy, tool distribution entropy, and trajectory distance scores than both the API-only and UI-only baselines across all settings, while also uncovering substantially more unique agent failures. These findings are consistent with prior observations from \citep{shea_sage_2025} that broader interaction diversity leads to improved failure discovery. Our metrics provide this same signal without requiring expensive LLM-judge-based analysis, suggesting that they can serve as an efficient proxy for estimating how effectively a simulator explores agent capabilities and failure modes.

\subsection{Case Study}
% \vspace{-0.4em}
% To further illustrate differences between the hybrid simulator and UI-only baseline, we present four interactions from our human evaluation (Figure \ref{fig:case_study}). The goal of our simulator is to aid evaluation by identifying product agent failures, therefore we focus specifically on these cases.

To illustrate how API access changes the behavior of the UI-capable simulator, we present four interactions comparing the hybrid and UI-only configurations under matched task scenarios (Figure~\ref{fig:case_study}). The goal of our simulator is to aid evaluation by identifying product agent failures; therefore, we focus specifically on cases in which the two simulators differ in their ability to expose such failures.

In the e-commerce example, the hybrid simulator responds to the product agent's request for an order number by directly stating the number without unnecessary additional information. This triggers an error in the product agent, which concludes that the order cannot be found instead of asking an appropriate follow-up question for account verification. In contrast, the UI-only simulator deviates from realistic user behavior by oversharing information about its account despite not being prompted for it. As a result, the shopping agent can complete the return request successfully but a realistic failure mode remains undiscovered. This issue stems from the fact that the UI-only simulator must navigate through several interface pages to retrieve its order information, including the login and account pages. During this process, the simulator accumulates additional account details in its context and unrealistically provides extra information the agent did not ask for, therefore missing the opportunity of detecting agent error like the API-based simulator.

A similar pattern emerges in the education domain. The hybrid simulator uses API-based actions to retrieve more precise account and test-launch information, allowing it to report an unexpected issue where the launch button is missing. This exposes a realistic limitation in the education agent, which falls back to generic troubleshooting steps rather than recognizing that the missing launch button may indicate a fulfillment or configuration problem. In contrast, the UI-only simulator repeats the problems already suggested by the education agent, in this case, reporting a blank screen. Since the interaction remains within the scope of the agent’s troubleshooting flow, the agent can provide a generic response and the failure mode remains unrevealed.

\section{Conclusion}
\label{sec:conclusion}

User simulation has emerged as a promising approach for scalable evaluation of interactive agents, but existing frameworks are often limited to either UI-based or API-based interactions and provide limited methods for evaluating simulator quality. To address these limitations, we introduced \framework{}, a flexible evaluation toolkit that combines a hybrid user simulator with a suite of metrics for measuring realism, capability coverage, and interaction effectiveness. Through experiments in e-commerce shopping and online education environments, we demonstrated that the hybrid simulator produces more realistic and comprehensive evaluations than UI-only approaches while uncovering a broader range of agent failure modes. We further showed that our proposed metrics align closely with human judgments and provide meaningful signals regarding simulator quality and evaluation coverage.

\section{Limitations}
\label{sec:limitations}

Although \framework{} improves the flexibility and comprehensiveness of simulation-based agent evaluation, several limitations remain. Like other LLM-based simulators, our framework does not perfectly model real human behavior. Recent studies have raised concerns about treating LLM-simulated users as human proxies. \citet{zhou_mind_2026} show that LLM simulators can be overly cooperative and stylistically uniform, while \citet{seshadri_lost_2026} find that simulated users may miscalibrate agent performance and surface different failure patterns than real humans. While our hybrid simulator increases interaction flexibility through combined UI and API actions, it may still exhibit behavioral biases inherited from the underlying LLMs.

Additionally, our evaluation is limited to two application domains: e-commerce shopping and online education support. Although these environments represent common customer-facing agent settings, additional experiments across other domains such as healthcare or enterprise productivity would help further validate the generality of the framework. Besides, our proposed metrics primarily measure coverage and interaction quality rather than directly estimating real-world deployment success. Future work could investigate stronger correlations between simulator-based metrics and downstream deployment outcomes.

Finally, hybrid interaction introduces additional system complexity. Coordinating between UI-based and API-based actions requires routing logic and environment-specific integrations, which may increase implementation overhead compared to simpler simulation setups. Despite these limitations, we believe \framework{} provides a useful step toward more flexible and comprehensive evaluation of modern interactive agents.

% \section{Acknowledgments}
% This research received funding from …. as well as DAPLab corporate support in the form of funding and/or compute from Amazon, IntellectAI, Infosys, Tidalwave, Veris, shopify, Microsoft, Thinking Machines, Dandy, Perplexity, and Daytona. The views and conclusions presented here are those of the authors and should not be interpreted as representing the official positions of the funding organizations.

% \section{Ethics Statement}

% \section{Acknowledgments}

\bibliography{custom}

@misc{zhu_realusersim_2026,
	title = {{RealUserSim}: {Bridging} the {Reality} {Gap} in {Agent} {Benchmarking} via {Grounded} {User} {Simulation}},
	shorttitle = {{RealUserSim}},
	url = {http://arxiv.org/abs/2605.20204},
	doi = {10.48550/arXiv.2605.20204},
	urldate = {2026-05-23},
	publisher = {arXiv},
	author = {Zhu, Ming and Tan, Juntao and Murthy, Rithesh and Qiu, Jielin and Yang, Liangwei and Zhao, Wenting and Savarese, Silvio and Heinecke, Shelby and Wang, Huan},
	month = apr,
	year = {2026},
	note = {arXiv:2605.20204 [cs.HC]},
}

@inproceedings{dong-etal-2025-large,
    title = "Large Language Model Agents in Finance: A Survey Bridging Research, Practice, and Real-World Deployment",
    author = "Dong, Yifei  and
      Wu, Fengyi  and
      Zhang, Kunlin  and
      Dai, Yilong  and
      Zhang, Sanjian  and
      Ye, Wanghao  and
      Chen, Sihan  and
      Cheng, Zhi-Qi",
    editor = "Christodoulopoulos, Christos  and
      Chakraborty, Tanmoy  and
      Rose, Carolyn  and
      Peng, Violet",
    booktitle = "Findings of the Association for Computational Linguistics: EMNLP 2025",
    month = nov,
    year = "2025",
    address = "Suzhou, China",
    publisher = "Association for Computational Linguistics",
    url = "https://aclanthology.org/2025.findings-emnlp.972/",
    doi = "10.18653/v1/2025.findings-emnlp.972",
    pages = "17889--17907",
    ISBN = "979-8-89176-335-7"
}

@misc{chu_llm_2026,
	title = {{LLM} {Agents} for {Education}: {Advances} and {Applications}},
	shorttitle = {{LLM} {Agents} for {Education}},
	url = {http://arxiv.org/abs/2503.11733},
	doi = {10.48550/arXiv.2503.11733},
	urldate = {2026-05-23},
	publisher = {arXiv},
	author = {Chu, Zhendong and Wang, Shen and Xie, Jian and Zhu, Tinghui and Yan, Yibo and Ye, Jinheng and Zhong, Aoxiao and Hu, Xuming and Liang, Jing and Yu, Philip S. and Wen, Qingsong},
	month = feb,
	year = {2026},
	note = {arXiv:2503.11733 [cs.CY]},
}

@misc{AmazonAlexaShopping2026,
  author = {{Amazon}},
  title = {Amazon Alexa},
  url = {https://amazon.com},
  year = {2026},
  month = {may},
  note = {Shopping via voice assistant}
}

@inproceedings{mohammadi_evaluation_2025,
	title = {Evaluation and {Benchmarking} of {LLM} {Agents}: {A} {Survey}},
	shorttitle = {Evaluation and {Benchmarking} of {LLM} {Agents}},
	url = {http://arxiv.org/abs/2507.21504},
	doi = {10.1145/3711896.3736570},
	urldate = {2026-05-23},
	booktitle = {Proceedings of the 31st {ACM} {SIGKDD} {Conference} on {Knowledge} {Discovery} and {Data} {Mining} {V}.2},
	author = {Mohammadi, Mahmoud and Li, Yipeng and Lo, Jane and Yip, Wendy},
	month = aug,
	year = {2025},
	note = {arXiv:2507.21504 [cs.LG]},
	pages = {6129--6139},
}

@misc{liu_agentbench_2025,
	title = {{AgentBench}: {Evaluating} {LLMs} as {Agents}},
	shorttitle = {{AgentBench}},
	url = {http://arxiv.org/abs/2308.03688},
	doi = {10.48550/arXiv.2308.03688},
	urldate = {2026-05-23},
	publisher = {arXiv},
	author = {Liu, Xiao and Yu, Hao and Zhang, Hanchen and Xu, Yifan and Lei, Xuanyu and Lai, Hanyu and Gu, Yu and Ding, Hangliang and Men, Kaiwen and Yang, Kejuan and Zhang, Shudan and Deng, Xiang and Zeng, Aohan and Du, Zhengxiao and Zhang, Chenhui and Shen, Sheng and Zhang, Tianjun and Su, Yu and Sun, Huan and Huang, Minlie and Dong, Yuxiao and Tang, Jie},
	month = oct,
	year = {2025},
	note = {arXiv:2308.03688 [cs.AI]},
}

@misc{zhou_webarena_2024,
	title = {{WebArena}: {A} {Realistic} {Web} {Environment} for {Building} {Autonomous} {Agents}},
	shorttitle = {{WebArena}},
	url = {http://arxiv.org/abs/2307.13854},
	doi = {10.48550/arXiv.2307.13854},
	urldate = {2026-05-23},
	publisher = {arXiv},
	author = {Zhou, Shuyan and Xu, Frank F. and Zhu, Hao and Zhou, Xuhui and Lo, Robert and Sridhar, Abishek and Cheng, Xianyi and Ou, Tianyue and Bisk, Yonatan and Fried, Daniel and Alon, Uri and Neubig, Graham},
	month = apr,
	year = {2024},
	note = {arXiv:2307.13854 [cs.AI]},
}

@misc{dou_simulatorarena_2025,
	title = {{SimulatorArena}: {Are} {User} {Simulators} {Reliable} {Proxies} for {Multi}-{Turn} {Evaluation} of {AI} {Assistants}?},
	shorttitle = {{SimulatorArena}},
	url = {http://arxiv.org/abs/2510.05444},
	doi = {10.48550/arXiv.2510.05444},
	urldate = {2026-05-23},
	publisher = {arXiv},
	author = {Dou, Yao and Galley, Michel and Peng, Baolin and Kedzie, Chris and Cai, Weixin and Ritter, Alan and Quirk, Chris and Xu, Wei and Gao, Jianfeng},
	month = oct,
	year = {2025},
	note = {arXiv:2510.05444 [cs.CL]},
}

@misc{wang_human_2025,
	title = {Human vs. {Agent} in {Task}-{Oriented} {Conversations}},
	url = {http://arxiv.org/abs/2509.17619},
	doi = {10.48550/arXiv.2509.17619},
	urldate = {2026-05-23},
	publisher = {arXiv},
	author = {Wang, Zhefan and Geng, Ning and Guo, Zhiqiang and Ma, Weizhi and Zhang, Min},
	month = nov,
	year = {2025},
	note = {arXiv:2509.17619 [cs.IR]},
}

@misc{yao_-bench_2024,
	title = {\$τ\$-bench: {A} {Benchmark} for {Tool}-{Agent}-{User} {Interaction} in {Real}-{World} {Domains}},
	shorttitle = {\$τ\$-bench},
	url = {http://arxiv.org/abs/2406.12045},
	urldate = {2024-11-08},
	publisher = {arXiv},
	author = {Yao, Shunyu and Shinn, Noah and Razavi, Pedram and Narasimhan, Karthik},
	month = jun,
	year = {2024},
	note = {arXiv:2406.12045},
}

@misc{huang_crmarena_2024,
	title = {{CRMArena}: {Understanding} the {Capacity} of {LLM} {Agents} to {Perform} {Professional} {CRM} {Tasks} in {Realistic} {Environments}},
	shorttitle = {{CRMArena}},
	url = {http://arxiv.org/abs/2411.02305},
	urldate = {2024-11-08},
	publisher = {arXiv},
	author = {Huang, Kung-Hsiang and Prabhakar, Akshara and Dhawan, Sidharth and Mao, Yixin and Wang, Huan and Savarese, Silvio and Xiong, Caiming and Laban, Philippe and Wu, Chien-Sheng},
	month = nov,
	year = {2024},
	note = {arXiv:2411.02305},
}

@misc{savadikar_shopgym_2026,
	title = {{ShopGym}: {An} {Integrated} {Framework} for {Realistic} {Simulation} and {Scalable} {Benchmarking} of {E}-{Commerce} {Web} {Agents}},
	shorttitle = {{ShopGym}},
	url = {http://arxiv.org/abs/2605.16116},
	doi = {10.48550/arXiv.2605.16116},
	urldate = {2026-05-23},
	publisher = {arXiv},
	author = {Savadikar, Chinmay and Zhao, Mingyu and Zhu, Yuanzheng and Li, Han and Xie, Shuang and Castelo, Alberto and Wu, Tianfu and Wang, Lingyun},
	month = may,
	year = {2026},
	note = {arXiv:2605.16116 [cs.AI]},
}

@misc{yehudai_survey_2026,
	title = {Survey on {Evaluation} of {LLM}-based {Agents}},
	url = {http://arxiv.org/abs/2503.16416},
	doi = {10.48550/arXiv.2503.16416},
	urldate = {2026-05-23},
	publisher = {arXiv},
	author = {Yehudai, Asaf and Eden, Lilach and Li, Alan and Uziel, Guy and Zhao, Yilun and Bar-Haim, Roy and Cohan, Arman and Shmueli-Scheuer, Michal},
	month = apr,
	year = {2026},
	note = {arXiv:2503.16416 [cs.AI]
version: 2},
}

@misc{castelo_simgym_2026,
	title = {{SimGym}: {Traffic}-{Grounded} {Browser} {Agents} for {Offline} {A}/{B} {Testing} in {E}-{Commerce}},
	shorttitle = {{SimGym}},
	url = {http://arxiv.org/abs/2602.01443},
	doi = {10.48550/arXiv.2602.01443},
	urldate = {2026-02-10},
	publisher = {arXiv},
	author = {Castelo, Alberto and Foumani, Zahra Zanjani and Fan, Ailin and Koay, Keat Yang and Malik, Vibhor and Zhu, Yuanzheng and Li, Han and Feghhi, Meysam and Uliana, Ronie and Xie, Shuang and Zhang, Zhaoyu and Martins, Angelo Ocana and Zhao, Mingyu and Pelland, Francis and Faerman, Jonathan and LeBlanc, Nikolas and Glazer, Aaron and McNamara, Andrew and Wang, Lingyun and Wu, Zhong},
	month = feb,
	year = {2026},
	note = {arXiv:2602.01443 [cs]},
}

@misc{sun_llm_2025,
	title = {{LLM} {Agent} {Meets} {Agentic} {AI}: {Can} {LLM} {Agents} {Simulate} {Customers} to {Evaluate} {Agentic}-{AI}-based {Shopping} {Assistants}?},
	shorttitle = {{LLM} {Agent} {Meets} {Agentic} {AI}},
	url = {http://arxiv.org/abs/2509.21501},
	doi = {10.48550/arXiv.2509.21501},
	urldate = {2025-09-30},
	publisher = {arXiv},
	author = {Sun, Lu and Fu, Shihan and Yao, Bingsheng and Lu, Yuxuan and Li, Wenbo and Gu, Hansu and Gesi, Jiri and Huang, Jing and Luo, Chen and Wang, Dakuo},
	month = sep,
	year = {2025},
	note = {arXiv:2509.21501 [cs]
version: 1},
}

@misc{shea_sage_2025,
	title = {{SAGE}: {A} {Top}-{Down} {Bottom}-{Up} {Knowledge}-{Grounded} {User} {Simulator} for {Multi}-turn {AGent} {Evaluation}},
	shorttitle = {{SAGE}},
	url = {http://arxiv.org/abs/2510.11997},
	doi = {10.48550/arXiv.2510.11997},
	urldate = {2026-01-04},
	publisher = {arXiv},
	author = {Shea, Ryan and Lu, Yunan and Qiu, Liang and Yu, Zhou},
	month = oct,
	year = {2025},
	note = {arXiv:2510.11997 [cs]},
}

@misc{zhu_where_2025,
	title = {Where {LLM} {Agents} {Fail} and {How} {They} can {Learn} {From} {Failures}},
	url = {http://arxiv.org/abs/2509.25370},
	doi = {10.48550/arXiv.2509.25370},
	urldate = {2026-01-06},
	publisher = {arXiv},
	author = {Zhu, Kunlun and Liu, Zijia and Li, Bingxuan and Tian, Muxin and Yang, Yingxuan and Zhang, Jiaxun and Han, Pengrui and Xie, Qipeng and Cui, Fuyang and Zhang, Weijia and Ma, Xiaoteng and Yu, Xiaodong and Ramesh, Gowtham and Wu, Jialian and Liu, Zicheng and Lu, Pan and Zou, James and You, Jiaxuan},
	month = sep,
	year = {2025},
	note = {arXiv:2509.25370 [cs]},
}

@misc{deng_mobile-bench_2024,
	title = {Mobile-{Bench}: {An} {Evaluation} {Benchmark} for {LLM}-based {Mobile} {Agents}},
	shorttitle = {Mobile-{Bench}},
	url = {http://arxiv.org/abs/2407.00993},
	doi = {10.48550/arXiv.2407.00993},
	urldate = {2026-03-11},
	publisher = {arXiv},
	author = {Deng, Shihan and Xu, Weikai and Sun, Hongda and Liu, Wei and Tan, Tao and Liu, Jianfeng and Li, Ang and Luan, Jian and Wang, Bin and Yan, Rui and Shang, Shuo},
	month = jul,
	year = {2024},
	note = {arXiv:2407.00993 [cs]},
}

@inproceedings{lu_uxagent_2025,
	title = {{UXAgent}: {An} {LLM} {Agent}-{Based} {Usability} {Testing} {Framework} for {Web} {Design}},
	shorttitle = {{UXAgent}},
	url = {http://arxiv.org/abs/2502.12561},
	doi = {10.1145/3706599.3719729},
	urldate = {2026-04-29},
	booktitle = {Proceedings of the {Extended} {Abstracts} of the {CHI} {Conference} on {Human} {Factors} in {Computing} {Systems}},
	author = {Lu, Yuxuan and Yao, Bingsheng and Gu, Hansu and Huang, Jing and Wang, Jessie and Li, Yang and Gesi, Jiri and He, Qi and Li, Toby Jia-Jun and Wang, Dakuo},
	month = apr,
	year = {2025},
	note = {arXiv:2502.12561 [cs]},
	pages = {1--12},
}

@misc{barres_2-bench_2025,
	title = {\$τ{\textasciicircum}2\$-{Bench}: {Evaluating} {Conversational} {Agents} in a {Dual}-{Control} {Environment}},
	shorttitle = {\$τ{\textasciicircum}2\$-{Bench}},
	url = {http://arxiv.org/abs/2506.07982},
	doi = {10.48550/arXiv.2506.07982},
	urldate = {2026-05-15},
	publisher = {arXiv},
	author = {Barres, Victor and Dong, Honghua and Ray, Soham and Si, Xujie and Narasimhan, Karthik},
	month = jun,
	year = {2025},
	note = {arXiv:2506.07982 [cs.AI]},
}

@misc{zhou_mind_2026,
	title = {Mind the {Sim2Real} {Gap} in {User} {Simulation} for {Agentic} {Tasks}},
	url = {http://arxiv.org/abs/2603.11245},
	doi = {10.48550/arXiv.2603.11245},
	urldate = {2026-05-20},
	publisher = {arXiv},
	author = {Zhou, Xuhui and Sun, Weiwei and Ma, Qianou and Xie, Yiqing and Liu, Jiarui and Du, Weihua and Welleck, Sean and Yang, Yiming and Neubig, Graham and Wu, Sherry Tongshuang and Sap, Maarten},
	month = mar,
	year = {2026},
	note = {arXiv:2603.11245 [cs.AI]},
}

@misc{seshadri_lost_2026,
	title = {Lost in {Simulation}: {LLM}-{Simulated} {Users} are {Unreliable} {Proxies} for {Human} {Users} in {Agentic} {Evaluations}},
	shorttitle = {Lost in {Simulation}},
	url = {http://arxiv.org/abs/2601.17087},
	doi = {10.48550/arXiv.2601.17087},
	urldate = {2026-05-20},
	publisher = {arXiv},
	author = {Seshadri, Preethi and Cahyawijaya, Samuel and Odumakinde, Ayomide and Singh, Sameer and Goldfarb-Tarrant, Seraphina},
	month = jan,
	year = {2026},
	note = {arXiv:2601.17087 [cs.HC]},
}

@misc{chopra_beyond_2026,
	title = {Beyond {Cooperative} {Simulators}: {Generating} {Realistic} {User} {Personas} for {Robust} {Evaluation} of {LLM} {Agents}},
	shorttitle = {Beyond {Cooperative} {Simulators}},
	url = {http://arxiv.org/abs/2605.12894},
	doi = {10.48550/arXiv.2605.12894},
	urldate = {2026-05-21},
	publisher = {arXiv},
	author = {Chopra, Harshita and Ghate, Kshitish and Caliskan, Aylin and Kohno, Tadayoshi and Shah, Chirag and Jaques, Natasha},
	month = may,
	year = {2026},
	note = {arXiv:2605.12894 [cs.AI]},
}

@misc{cemri_why_2025,
	title = {Why {Do} {Multi}-{Agent} {LLM} {Systems} {Fail}?},
	url = {http://arxiv.org/abs/2503.13657},
	doi = {10.48550/arXiv.2503.13657},
	urldate = {2026-05-22},
	publisher = {arXiv},
	author = {Cemri, Mert and Pan, Melissa Z. and Yang, Shuyi and Agrawal, Lakshya A. and Chopra, Bhavya and Tiwari, Rishabh and Keutzer, Kurt and Parameswaran, Aditya and Klein, Dan and Ramchandran, Kannan and Zaharia, Matei and Gonzalez, Joseph E. and Stoica, Ion},
	month = oct,
	year = {2025},
	note = {arXiv:2503.13657 [cs.AI]},
}

@misc{singh2026openaigpt5card,
      title={OpenAI GPT-5 System Card}, 
      author={Aaditya Singh and Adam Fry and Adam Perelman and Adam Tart and Adi Ganesh and Ahmed El-Kishky and Aidan McLaughlin and Aiden Low and AJ Ostrow and Akhila Ananthram and Akshay Nathan and Alan Luo and Alec Helyar and Aleksander Madry and Aleksandr Efremov and Aleksandra Spyra and Alex Baker-Whitcomb and Alex Beutel and Alex Karpenko and Alex Makelov and Alex Neitz and Alex Wei and Alexandra Barr and Alexandre Kirchmeyer and Alexey Ivanov and Alexi Christakis and Alistair Gillespie and Allison Tam and Ally Bennett and Alvin Wan and Alyssa Huang and Amy McDonald Sandjideh and Amy Yang and Ananya Kumar and Andre Saraiva and Andrea Vallone and Andrei Gheorghe and Andres Garcia Garcia and Andrew Braunstein and Andrew Liu and Andrew Schmidt and Andrey Mereskin and Andrey Mishchenko and Andy Applebaum and Andy Rogerson and Ann Rajan and Annie Wei and Anoop Kotha and Anubha Srivastava and Anushree Agrawal and Arun Vijayvergiya and Ashley Tyra and Ashvin Nair and Avi Nayak and Ben Eggers and Bessie Ji and Beth Hoover and Bill Chen and Blair Chen and Boaz Barak and Borys Minaiev and Botao Hao and Bowen Baker and Brad Lightcap and Brandon McKinzie and Brandon Wang and Brendan Quinn and Brian Fioca and Brian Hsu and Brian Yang and Brian Yu and Brian Zhang and Brittany Brenner and Callie Riggins Zetino and Cameron Raymond and Camillo Lugaresi and Carolina Paz and Cary Hudson and Cedric Whitney and Chak Li and Charles Chen and Charlotte Cole and Chelsea Voss and Chen Ding and Chen Shen and Chengdu Huang and Chris Colby and Chris Hallacy and Chris Koch and Chris Lu and Christina Kaplan and Christina Kim and CJ Minott-Henriques and Cliff Frey and Cody Yu and Coley Czarnecki and Colin Reid and Colin Wei and Cory Decareaux and Cristina Scheau and Cyril Zhang and Cyrus Forbes and Da Tang and Dakota Goldberg and Dan Roberts and Dana Palmie and Daniel Kappler and Daniel Levine and Daniel Wright and Dave Leo and David Lin and David Robinson and Declan Grabb and Derek Chen and Derek Lim and Derek Salama and Dibya Bhattacharjee and Dimitris Tsipras and Dinghua Li and Dingli Yu and DJ Strouse and Drew Williams and Dylan Hunn and Ed Bayes and Edwin Arbus and Ekin Akyurek and Elaine Ya Le and Elana Widmann and Eli Yani and Elizabeth Proehl and Enis Sert and Enoch Cheung and Eri Schwartz and Eric Han and Eric Jiang and Eric Mitchell and Eric Sigler and Eric Wallace and Erik Ritter and Erin Kavanaugh and Evan Mays and Evgenii Nikishin and Fangyuan Li and Felipe Petroski Such and Filipe de Avila Belbute Peres and Filippo Raso and Florent Bekerman and Foivos Tsimpourlas and Fotis Chantzis and Francis Song and Francis Zhang and Gaby Raila and Garrett McGrath and Gary Briggs and Gary Yang and Giambattista Parascandolo and Gildas Chabot and Grace Kim and Grace Zhao and Gregory Valiant and Guillaume Leclerc and Hadi Salman and Hanson Wang and Hao Sheng and Haoming Jiang and Haoyu Wang and Haozhun Jin and Harshit Sikchi and Heather Schmidt and Henry Aspegren and Honglin Chen and Huida Qiu and Hunter Lightman and Ian Covert and Ian Kivlichan and Ian Silber and Ian Sohl and Ibrahim Hammoud and Ignasi Clavera and Ikai Lan and Ilge Akkaya and Ilya Kostrikov and Irina Kofman and Isak Etinger and Ishaan Singal and Jackie Hehir and Jacob Huh and Jacqueline Pan and Jake Wilczynski and Jakub Pachocki and James Lee and James Quinn and Jamie Kiros and Janvi Kalra and Jasmyn Samaroo and Jason Wang and Jason Wolfe and Jay Chen and Jay Wang and Jean Harb and Jeffrey Han and Jeffrey Wang and Jennifer Zhao and Jeremy Chen and Jerene Yang and Jerry Tworek and Jesse Chand and Jessica Landon and Jessica Liang and Ji Lin and Jiancheng Liu and Jianfeng Wang and Jie Tang and Jihan Yin and Joanne Jang and Joel Morris and Joey Flynn and Johannes Ferstad and Johannes Heidecke and John Fishbein and John Hallman and Jonah Grant and Jonathan Chien and Jonathan Gordon and Jongsoo Park and Jordan Liss and Jos Kraaijeveld and Joseph Guay and Joseph Mo and Josh Lawson and Josh McGrath and Joshua Vendrow and Joy Jiao and Julian Lee and Julie Steele and Julie Wang and Junhua Mao and Kai Chen and Kai Hayashi and Kai Xiao and Kamyar Salahi and Kan Wu and Karan Sekhri and Karan Sharma and Karan Singhal and Karen Li and Kenny Nguyen and Keren Gu-Lemberg and Kevin King and Kevin Liu and Kevin Stone and Kevin Yu and Kristen Ying and Kristian Georgiev and Kristie Lim and Kushal Tirumala and Kyle Miller and Lama Ahmad and Larry Lv and Laura Clare and Laurance Fauconnet and Lauren Itow and Lauren Yang and Laurentia Romaniuk and Leah Anise and Lee Byron and Leher Pathak and Leon Maksin and Leyan Lo and Leyton Ho and Li Jing and Liang Wu and Liang Xiong and Lien Mamitsuka and Lin Yang and Lindsay McCallum and Lindsey Held and Liz Bourgeois and Logan Engstrom and Lorenz Kuhn and Louis Feuvrier and Lu Zhang and Lucas Switzer and Lukas Kondraciuk and Lukasz Kaiser and Manas Joglekar and Mandeep Singh and Mandip Shah and Manuka Stratta and Marcus Williams and Mark Chen and Mark Sun and Marselus Cayton and Martin Li and Marvin Zhang and Marwan Aljubeh and Matt Nichols and Matthew Haines and Max Schwarzer and Mayank Gupta and Meghan Shah and Melody Y. Guan and Melody Huang and Meng Dong and Mengqing Wang and Mia Glaese and Micah Carroll and Michael Lampe and Michael Malek and Michael Sharman and Michael Zhang and Michele Wang and Michelle Pokrass and Mihai Florian and Mikhail Pavlov and Miles Wang and Ming Chen and Mingxuan Wang and Minnia Feng and Mo Bavarian and Molly Lin and Moose Abdool and Mostafa Rohaninejad and Nacho Soto and Natalie Staudacher and Natan LaFontaine and Nathan Marwell and Nelson Liu and Nick Preston and Nick Turley and Nicklas Ansman and Nicole Blades and Nikil Pancha and Nikita Mikhaylin and Niko Felix and Nikunj Handa and Nishant Rai and Nitish Keskar and Noam Brown and Ofir Nachum and Oleg Boiko and Oleg Murk and Olivia Watkins and Oona Gleeson and Pamela Mishkin and Patryk Lesiewicz and Paul Baltescu and Pavel Belov and Peter Zhokhov and Philip Pronin and Phillip Guo and Phoebe Thacker and Qi Liu and Qiming Yuan and Qinghua Liu and Rachel Dias and Rachel Puckett and Rahul Arora and Ravi Teja Mullapudi and Raz Gaon and Reah Miyara and Rennie Song and Rishabh Aggarwal and RJ Marsan and Robel Yemiru and Robert Xiong and Rohan Kshirsagar and Rohan Nuttall and Roman Tsiupa and Ronen Eldan and Rose Wang and Roshan James and Roy Ziv and Rui Shu and Ruslan Nigmatullin and Saachi Jain and Saam Talaie and Sam Altman and Sam Arnesen and Sam Toizer and Sam Toyer and Samuel Miserendino and Sandhini Agarwal and Sarah Yoo and Savannah Heon and Scott Ethersmith and Sean Grove and Sean Taylor and Sebastien Bubeck and Sever Banesiu and Shaokyi Amdo and Shengjia Zhao and Sherwin Wu and Shibani Santurkar and Shiyu Zhao and Shraman Ray Chaudhuri and Shreyas Krishnaswamy and Shuaiqi and Xia and Shuyang Cheng and Shyamal Anadkat and Simón Posada Fishman and Simon Tobin and Siyuan Fu and Somay Jain and Song Mei and Sonya Egoian and Spencer Kim and Spug Golden and SQ Mah and Steph Lin and Stephen Imm and Steve Sharpe and Steve Yadlowsky and Sulman Choudhry and Sungwon Eum and Suvansh Sanjeev and Tabarak Khan and Tal Stramer and Tao Wang and Tao Xin and Tarun Gogineni and Taya Christianson and Ted Sanders and Tejal Patwardhan and Thomas Degry and Thomas Shadwell and Tianfu Fu and Tianshi Gao and Timur Garipov and Tina Sriskandarajah and Toki Sherbakov and Tomek Korbak and Tomer Kaftan and Tomo Hiratsuka and Tongzhou Wang and Tony Song and Tony Zhao and Troy Peterson and Val Kharitonov and Victoria Chernova and Vineet Kosaraju and Vishal Kuo and Vitchyr Pong and Vivek Verma and Vlad Petrov and Wanning Jiang and Weixing Zhang and Wenda Zhou and Wenlei Xie and Wenting Zhan and Wes McCabe and Will DePue and Will Ellsworth and Wulfie Bain and Wyatt Thompson and Xiangning Chen and Xiangyu Qi and Xin Xiang and Xinwei Shi and Yann Dubois and Yaodong Yu and Yara Khakbaz and Yifan Wu and Yilei Qian and Yin Tat Lee and Yinbo Chen and Yizhen Zhang and Yizhong Xiong and Yonglong Tian and Young Cha and Yu Bai and Yu Yang and Yuan Yuan and Yuanzhi Li and Yufeng Zhang and Yuguang Yang and Yujia Jin and Yun Jiang and Yunyun Wang and Yushi Wang and Yutian Liu and Zach Stubenvoll and Zehao Dou and Zheng Wu and Zhigang Wang},
      year={2026},
      eprint={2601.03267},
      archivePrefix={arXiv},
      primaryClass={cs.CL},
      url={https://arxiv.org/abs/2601.03267}, 
}

@misc{qwenteam2026qwen35omnitechnicalreport,
      title={Qwen3.5-Omni Technical Report}, 
      author={{Qwen Team}},
      year={2026},
      eprint={2604.15804},
      archivePrefix={arXiv},
      primaryClass={cs.CL},
      url={https://arxiv.org/abs/2604.15804}, 
}

\clearpage
\appendix
\newpage

\section{Human Evaluation Details}
\label{sec:app_human_eval}
\subsection{}
\label{sec:app_human_eval1}

To assess the realism of the generated user messages, we recruited two graduate students as annotators. Both annotators had prior experience developing interactive agents, which qualified them to perform the task. For each use case, annotators evaluated user utterances from 150 conversations along three dimensions: \textit{human-likeness}, \textit{coherence}, and \textit{goal consistency}. Each dimension was rated on a 3-point Likert scale. Inter-annotator agreement, measured by Cohen's kappa, is 0.45 on average across the three dimensions, indicating moderate agreement. Additionally, we use Kendall's $\tau$ to measure the consistency of annotators' relative rankings. We observe statistically significant positive correlations across all three dimensions ($p < 0.05$): human-likeness achieves $\tau=0.661$ , coherence $\tau=0.592$, and goal consistency $\tau=0.473$. These results indicate moderate to strong agreement, suggesting that while annotators may occasionally differ by one rating level, they generally rank conversations consistently. The detailed annotation guidelines are provided below. Our data collection protocol was approved by the Institutional Review Board (IRB).

\begin{tcolorbox}[
  breakable,
  enhanced,
  colback=gray!5!white,
  colframe=gray!60!black,
  % boxsep=1pt, 
  left=5pt, 
  right=5pt, 
  % top=1pt, 
  % bottom=1pt
]
\small{
\textbf{1. Human-Likeness} \\
\textbf{Definition}: How closely the simulated user’s language resembles that of a human. \\
\textbf{Scale (1–3)}: \\
1 = Clearly artificial, awkward, or robotic phrasing. \\
2 = Somewhat human-like with occasional awkwardness or unnatural phrasing. \\
3 = Highly natural; indistinguishable from a real user. \\
\\
\textbf{2. Coherence} \\
\textbf{Definition}: How well the simulated user’s utterances follow logically from the conversation history. \\
\textbf{Scale (1–3)}: \\
1 = Often off-topic or logically inconsistent. \\
2 = Mostly coherent with occasional minor lapses. \\
3 = Fully coherent throughout. \\
\\
\textbf{3. Goal Consistency} \\
\textbf{Definition}: Whether the simulated user’s utterances remain consistent with the provided user goal. \\
\textbf{Scale (1–3)}: \\
1 = Frequently contradicts, ignores, or drifts away from the provided goal. \\
2 = Mostly follows the goal, but includes minor inconsistencies, omissions, or unnecessary deviations. \\
3 = Fully consistent with the provided goal across the conversation.
}
\end{tcolorbox}

\subsection{}
\label{sec:app_human_eval2}

Following prior work \cite{chopra_beyond_2026}, we conducted a blinded human evaluation on conversations from the education agent. Each conversation involved the same education agent interacting with a user from one of four sources: a real customer, the API-only simulator, the UI-only simulator, or our hybrid simulator. Both simulators used \texttt{gpt-5.4} as the backbone model.

We recruited four graduate students as annotators. Annotators were asked to review only the user-side messages in each conversation and determine whether they appeared to be written by a real human, generated by a bot, or too ambiguous to classify confidently. The evaluation included 300 conversations in total, with 75 conversations from each user source.

\section{Education Agent User Tools}
\label{sec:app_edu_tools}

We create eight mocked user tools to support simulation-based evaluation of the education agent, as summarized in Table \ref{tab:edu_user_tools}. These tools model user-side actions that commonly arise in online education support scenarios, such as logging into an account, checking order records, rescheduling or canceling a test, contacting support, and launching a practice exam. Each tool returns a structured response with a status field, returned data, and an optional user-facing error message if action failed. Figure \ref{fig:mock_reschedule_api} shows one representative implementation for the rescheduling tool. This design allows the simulator to report realistic execution outcomes after following the agent's instructions, without accessing production systems or manually configuring live test accounts.

\begin{table}[t]
\centering
\small
\begin{tabular}{p{0.3\columnwidth}p{0.6\columnwidth}}
\toprule
\textbf{Failure Type} & \textbf{Description} \\
\midrule
\texttt{false information} 
& Contradicts with the ground truth knowledge. \\

\midrule
\texttt{disobey user request} 
& Completely ignores what the user asked. \\

\midrule
\texttt{lack of specific information} 
& Correct but incomplete answer that omits details present in the knowledge. \\

\midrule
\texttt{repetition} 
& Restates previous content without adding value. \\

\midrule
\texttt{failure to ask for clarification} 
& Proceeds with an assumption instead of asking the user for clarification. \\

\bottomrule
\end{tabular}
\caption{Agent Failure Types}
\label{tab:error_types}
\end{table}

\begin{table}[t]
\centering
\small
\begin{tabular}{p{0.32\columnwidth}p{0.62\columnwidth}}
\toprule
\textbf{User Tool} & \textbf{Description} \\
\midrule
\texttt{get\_account\_info} 
& Looks up a user's account record by email address from user database and returns profile, location, testing-region information, etc. \\

\midrule
\texttt{get\_order\_info} 
& Retrieves all orders associated with a given customer ID from user database, including order records and payment-related information. \\

\midrule
\texttt{login} 
& Simulates a login attempt and returns login outcomes, such as wrong password, account lock, reset-code failure, etc. \\

\midrule
\texttt{edit\_account} 
& Simulates an account-editing attempt and returns outcomes such as blocked email changes, name field uneditable, duplicate-account conflicts, etc. \\

\midrule
\texttt{contact\_support} 
& Simulates contacting support and returns realistic support outcomes, such as no response, a bounced email, etc. \\

\midrule
\texttt{reschedule\_test} 
& Simulates a test-rescheduling attempt and returns outcomes such as rescheduling not allowed, test already started, blocked mode switching, etc. \\

\midrule
\texttt{cancel\_test} 
& Simulates a test-cancellation attempt and returns outcomes such as cancellation not allowed, refund unavailable, etc. \\

\midrule
\texttt{launch\_exam} 
& Simulates launching an exam and returns outcomes such as missing launch buttons, questions not loading, access being blocked, etc. \\

\bottomrule
\end{tabular}
\caption{Mock API-based user tools used to build the hybrid user simulator for the education agent.}
\label{tab:edu_user_tools}
\end{table}

\begin{figure}[t]
\centering
\begin{minipage}{0.9\columnwidth}
\begin{lstlisting}[style=compactpython, numbers=none]
@tool
def reschedule_test(user_id, user_intent=None):
    """
    Simulate a user's attempt to reschedule an exam.
    Return either success or a realistic error.
    """

    user = get_account_info(user_id)
    order = get_order_info(user_id)

    if user is None or order is None:
        return {
            status: "failed",
            data: {reschedule_result: "record_not_found"},
            message: "No active exam was found."
        }

    related = Check(user_intent,
                    target_intent="reschedule test")

    # Return success when rescheduling is only an
    # intermediate step in another user task.
    if not related:
        return {
            status: "success",
            data: {reschedule_result: "success"},
            message: "No rescheduling issue."
        }

    possible_failures = [
        {
            status: "blocked",
            data: {reschedule_result: "not_allowed"},
            message: "Reschedule button is missing."
        },
        {
            status: "blocked",
            data: {reschedule_result: "already_started"},
            message: "The test has already started."
        },
        {
            status: "blocked",
            data: {reschedule_result: "mode_switch_blocked"},
            message: "Cannot switch test mode."
        },
        {
            status: "partial_success",
            data: {reschedule_result: "no_slots_available"},
            message: "No slots are available."
        },
        {
            status: "technical_error",
            data: {reschedule_result: "technical_error"},
            message: "The rescheduling page errored out."
        }
    ]

    return random_choice(possible_failures)
\end{lstlisting}
\end{minipage}
\caption{
Example mocked API for rescheduling tool. Given a user identifier and intent, the tool retrieves mocked account and order records, checks whether the intent concerns rescheduling, and samples a realistic platform outcome. The structured response allows the simulator to report diverse user-side execution results without accessing production systems.
}
\label{fig:mock_reschedule_api}
\end{figure}

\section{The Agent Failures Analysis}
\label{sec:app_agent_errors}
We aggregate the agent failure types following prior work on agent failure analysis \cite{shea_sage_2025, cemri_why_2025, zhu_where_2025}. Table \ref{tab:error_types} summarizes the resulting taxonomy, and Table \ref{tab:sample_errors} provides representative interaction examples for each failure type. Among these categories, the most severe failures are concentrated in \textit{false information} and \textit{disobeying user requests}. For false information, we observe cases where the product agent fabricates product links, either generating non-existent URLs based on product titles (e.g., \texttt{www.<product\_title>.com}) or providing links outside the agent’s knowledge base, such as Amazon product pages. For disobeying user requests, one failure stems from a system-level issue. In the middle of the conversation, the product agent reverts to its initial state after a failed tool call, ignores the user’s prior context, and repeats the greeting message instead of continuing the task.

\begin{table*}[t]
    \centering
    \small
    \begin{tabularx}{\textwidth}{X}
   
    \toprule
      \arrayrulecolor{gray!20}\specialrule{0pt}{0pt}{-2pt}
      \rowcolor{gray!20}\normalsize\textbf{false information} \\
      \arrayrulecolor{black}
    \toprule
      \textbf{Failure Description}: Claims to have performed restock notifications that the assistant cannot actually do. \\
      \midrule
      \textbf{Example}: \\
      \textbf{User}: When will the Nordic Simple Bedding Set Adult Duvet Cover Sets Bedclothes Bed Linen Sheet Single Double Queen King size Qulit Covers 240/220 be restocked? Please notify me at davidmiller0203@dcpa.net when it's back in stock. \\
      \textbf{Agent}: \textcolor{orange}{I've subscribed you to restock notifications for the Nordic Simple Bedding Set. You'll be notified at davidmiller0203@dcpa.net when it's back in stock.} \\

    \toprule
      \arrayrulecolor{gray!20}\specialrule{0pt}{0pt}{-2pt}
      \rowcolor{gray!20} \normalsize\textbf{disobey user request}\\
      \arrayrulecolor{black}
    \toprule
      \textbf{Failure Description}: Assistant asks for test location instead of performing the specifically requested link or policy. \\
      \midrule
      \textbf{User Goal}: Want to launch the practice test but got a issue. \\
      \midrule
      \textbf{Example}: \\
      \textit{[Earlier parts omitted for brevity]} \\
      \textbf{User}:  When I try to start the Social Studies practice test, it will not launch from my account. \\
      \textbf{Agent}: For issues with launching or accessing the test, the next step is to contact support team directly so they can look at your account...[A long suggestion about how to contact support team]\\
      \textbf{User}: Can you make it shorter so I can send it to support team? \\
      \textbf{Agent}: \textcolor{orange}{Could you please specify the country or US state where you plan to take the test? The answer can be different based on the location.}\\

    \toprule
      \arrayrulecolor{gray!20}\specialrule{0pt}{0pt}{-2pt}
      \rowcolor{gray!20}\normalsize\textbf{lack of specific information} \\
      \arrayrulecolor{black}
    \toprule
      \textbf{Failure Description}: Fails to give a specific contact methods for customer service \\
      \midrule
      \textbf{Example}: \\
      \textbf{User}: Do you offer digital gift cards? \\
      \textbf{Agent}: Yes, we offer digital gift cards that you can purchase directly from our website. \\
      \textbf{User}: Please give me the best contact method for support about digital gift card policy questions \\
      \textbf{Agent}: \textcolor{orange}{For questions regarding digital gift card policies, please reach out to our customer support team!} \\

    \toprule
      \arrayrulecolor{gray!20}\specialrule{0pt}{0pt}{-2pt}
      \rowcolor{gray!20}\normalsize\textbf{repetition} \\
      \arrayrulecolor{black}
    \toprule
      \textbf{Failure Description}: Repeating the same generic sustainability statements across turns without adding other values or provide alternative support solution. \\
      \midrule
      \textbf{Example}: \\
      \textbf{User}: What are your sustainability and eco-friendly product policies? \\
      \textbf{Agent}: We strive to use sustainable materials and partner with manufacturers who follow ethical production practices. Each product page provides detailed information about the materials used, prioritizing quality fabrics like cotton, polyester, and eco-friendly options when available. \\
      \textbf{User}: Do you have details on product materials, certifications, and policy link to look through? \\
      \textbf{Agent}: \textcolor{orange}{Yes, we strive to use sustainable materials and partner with manufacturers who prioritize ethical production practices. You can find detailed information about the materials used in each product on their respective pages, including options like cotton and eco-friendly fabrics when available.} \\
      \textit{[Later interaction repeat the same answer multiple times]} \\

    \toprule
      \arrayrulecolor{gray!20}\specialrule{0pt}{0pt}{-2pt}
      \rowcolor{gray!20}\normalsize\textbf{failure to ask for clarification} \\
      \arrayrulecolor{black}
    \toprule
      \textbf{Failure Description}: Fails to ask targeted clarification questions for a missing launch button, such as account, order fulfillment, and instead keeps giving generic browser troubleshooting after the user indicates the issue is not browser-related. \\
      \midrule
      \textbf{Example}: \\
      \textbf{User}: I have trouble launching my Social Studies practice test. \\
      \textit{[Earlier parts omitted for brevity]} \\
      \textbf{User}: There's no launch button for it. \\
      \textbf{Agent}: \textcolor{orange}{I do not have specific information about missing launch buttons. Here are some general steps for practice-test launch issues. 1. Try tech/browser fixes first. Use a computer or laptop if possible. Switch browser...} \\
      \textbf{User}: I tried those steps, but it is not a browser issue. I still cannot see the launch button. \\
      \textbf{Agent}:  \textcolor{orange}{Since you have already tried the account and browser steps, this is likely something we needs to fix. Please contact support team at https://example.com, submit contact form and describe your issue...} \\
    \bottomrule
    \end{tabularx}
    \caption{The synthesized interaction examples generated by the hybrid user simulator. The sentences highlighted in \textcolor{orange}{orange} represent the agent errors for the corresponding error category.}
    \label{tab:sample_errors}
\end{table*}

\clearpage
\onecolumn
\section{Prompts}

\subsection{User Agent}
\label{sec:app_prompt_user_agent}

\begin{tcolorbox}[
  breakable,
  enhanced,
  colback=gray!5!white,
  colframe=gray!60!black,
  title= Hybrid User Simulation Router,
]
\small{
\begin{lstlisting}[
  basicstyle=\ttfamily\small,
  breaklines=true,
  breakautoindent=false,
  breakindent=0pt,
  postbreak=
]
You are deciding the next step for a simulated user. Specifically, determine whether the simulated user should take an API-based action (e.g. calling a tool to gather more information) before taking a UI-based action (e.g. clicking a button or typing in the chat). Note that the user will always interact with the chat widget via UI actions eventually, but this decision is about whether a preliminary API call would help gather enough information to produce a more specific and relevant UI response.

The user's goal is:
{intent}

Tools list:
{tools_list} # The available user tools list with description

# Key Guidelines 

Choose "yes" if any of the listed API actions are relevant to the user's goal; otherwise choose "no".
Prefer API actions when in doubt.
However, if the conversation memory shows that a tool was already called but the chatbot has not yet responded or acted on the result, calling that same tool again will not produce a different outcome - return "no" in that case.

Return JSON with:
{
    "reasoning": "<brief explanation>",
    "decision": "yes" (take an API action first) or "no" (proceed directly with a UI action)
}

\end{lstlisting}
}
\end{tcolorbox}

\begin{tcolorbox}[
  breakable,
  enhanced,
  colback=gray!5!white,
  colframe=gray!60!black,
  title=System prompt for action module,
]
\small{
\begin{lstlisting}[
  basicstyle=\ttfamily\small,
  breaklines=true,
  breakautoindent=false,
  breakindent=0pt,
  postbreak=
]
You are interacting with a customer support chatbot embedded in a web page. You will be given the current HTML and your task is to translate the next step (of natural language string format) into a JSON actionable step defined in the action space.

## The chatbot is supplied by the following company:
{corporate_overview}

You are tasked with determining the **next action** based on the provided `persona`, `intent`, `plan`, `next_step`, and `environment`. Your job is to interpret `next_step` and translate it into **one** actionable step within the web environment.

---

# Input Explanation

1. **Persona**: Who I am; can influence decisions. Example: "I'm a PhD student with a stipend of $50k."
2. **Intent**: What I aim to achieve. Example: "I need help with my account."
3. **Plan**: A sequence of steps that leads to the intent.
4. **Next Step**: The specific step to execute now. Example: "Select the desired size 'Small'."
5. **Environment**: A structured snapshot that includes:
   - The current page HTML (string)
   - A list of input fields by their `data-semantic-id` attributes
   - A list of clickable elements by their `data-semantic-id` attributes

Use the `environment` lists to map actions to element **targets**.

---

# Key Guidelines

1. **Think in the first person.** I perform actions step-by-step from persona, intent, plan, and environment.
2. **Translate `next_step` into a concrete UI action.** Choose the correct element by matching its `data-semantic-id` to the action's `target`.
3. **Return exactly one action** in `{ "actions": [ { ... } ] }`. The array **must contain one and only one** action object. No extra text.
4. **Use `terminate` only if the step explicitly says to stop** (e.g., "terminate", "quit", "stop"). Otherwise produce an actionable step.
5. **Do not repeat a previous action.** Even if the prior attempt failed, never output the same action again.
6. **Compress compound interactions into a single action when applicable.** For example, typing a message in the chatbot input and sending it is a single `type` action with `"enter": true`. The chatbot input field is the primary target for `type` actions.
7. **Field naming:** UI elements are addressed with the field **`target`** (not `data-semantic-id`).
8. If multiple candidate targets exist, prefer an **exact match**; otherwise choose the closest normalized match (case-insensitive; ignore spaces/underscores/punctuation). **Never invent targets** not present in the environment.

---

# Action Space

Each action object must include an `action` key specifying one of the following types. Include required fields exactly as shown.

## Click
{ "action": "click", "target": "<element_semantic_id>", "description": "Clicking ..." }

## Type (with optional Enter submit)
{ "action": "type", "target": "<input_semantic_id>", "text": "<text>", "enter": true, "description": "Typing and submitting ..." }

## Hover
{ "action": "hover", "target": "<element_semantic_id>", "description": "Hovering ..." }

## Select (e.g., dropdowns)
{ "action": "select", "target": "<select_semantic_id>", "value": "<option_value>", "description": "Selecting ..." }

## Clear (clear an input field)
{ "action": "clear", "target": "<input_semantic_id>", "description": "Clearing ..." }

## Key Press (optionally scoped to a target)
{ "action": "key_press", "key": "Enter", "target": "<optional_element_semantic_id>", "description": "Pressing Enter ..." }

## Scroll (scroll the chat window or current page)
{ "action": "scroll", "target": "<optional_element_semantic_id>", "direction": "up", "amount": 300, "description": "Scrolling up ..." }
{ "action": "scroll", "target": "<optional_element_semantic_id>", "direction": "down", "amount": 300, "description": "Scrolling down ..." }

## Navigation (use when the chatbot provides a URL to open)
{ "action": "goto_url", "url": "https://example.com", "description": "Navigating ..." }
{ "action": "back",     "description": "Going back ..." }
{ "action": "forward",  "description": "Going forward ..." }
{ "action": "refresh",  "description": "Refreshing ..." }

## Tabs (use when a chatbot link opens in a new tab)
{ "action": "new_tab",    "url": "https://example.com", "description": "Opening in new tab ..." }
{ "action": "switch_tab", "tab_id": 1, "description": "Switching tab ..." }
{ "action": "close_tab",  "tab_id": 1, "description": "Closing tab ..." }

## Terminate (only if explicitly instructed by the step)
{ "action": "terminate", "description": "Terminating ..." }

---

# Element Targeting Rules

- Match `target` to an element `data-semantic-id` from the environment's input or clickable lists.
- Chatbot window elements (chat input, send button, product links/buttons in the chat) are the most common targets.
- You can *only* target input elements with the `type` action.
- `target` is optional for `scroll` - omit it to scroll the full page, or provide a container's `data-semantic-id` to scroll within that element.
- If multiple candidates exist, prefer exact match; otherwise choose the closest normalized match (case-insensitive, spaces/underscores/punctuation treated equivalently).
- Never invent targets not present in the environment snapshot.

---

# Critical Anti-Repeat Rules

- **No matter whether the previous action was successful or not, DO NOT REPEAT A PREVIOUS ACTION.**
- **Do not ask the same question again if it was already asked by a previous action.**
- **Do not return terminate actions unless your intent has been fulfilled.**
- **If you are waiting for a response after typing a message, then return {"actions": [{ "action": "scroll", "target": "<optional_element_semantic_id>", "direction": "down", "amount": 300, "description": "Scrolling down ..." }]} to wait for the response.**

---

# Output Format

Return **only** a single valid JSON object:

{
  "actions": [ { /* exactly one action object */ } ]
}

No extra text, no markdown, no commentary.
\end{lstlisting}
}
\end{tcolorbox}

\begin{figure}[H]
\centering
\begin{minipage}{0.95\columnwidth}
\begin{lstlisting}[style=compactpython, numbers=none]
user_payload = {
    "valid_targets": { # The interactive selectors from environment
        "inputs": inputs,
        "clickable": clickables,
        "selects": selects,
        "hoverable": hoverables,
        "tabs": tabs,
    },
    "persona": self.persona,
    "intent": self.intent,
    "plan": self.current_plan.content,
    "next_step": self.current_plan.next_step,
    "environment": env["html"],
    "tool_execution_information": tool_info, # If there is user tools response, then include this field.
}

messages = [
    {
        "role": "system",
        "content": system_prompt,
    },
    {
        "role": "user",
        "content": json.dumps(user_payload),
    },
]
\end{lstlisting}
\end{minipage}
\caption{Construction of the runtime messages for action module.}
\label{fig:runtime_user_prompt}
\end{figure}

\subsection{Realism Evaluator}
\label{sec:app_prompt_realism}

\begin{tcolorbox}[
  breakable,
  enhanced,
  colback=gray!5!white,
  colframe=gray!60!black,
  title= Goal Consistency,
]
\small{
\begin{lstlisting}[
  basicstyle=\ttfamily\small,
  breaklines=true,
  breakautoindent=false,
  breakindent=0pt,
  postbreak=
]
You are evaluating a user simulator in a conversation with an AI assistant.

You will be given:
- The user's stated goal
- The full conversation history up to a specific turn

Your task is to judge whether the user's messages in this conversation are consistent with their
stated goal. A user is consistent if their questions, requests, and responses logically pursue
the stated goal without contradicting or abandoning it.

Output ONLY valid JSON. No markdown, no extra text. Use this format exactly:
{
    "result": <true or false>,
    "reason": "<brief reasoning>"
}

User goal:
{goal}

Conversation history:
{conversation}

Return:

\end{lstlisting}
}
\end{tcolorbox}

\begin{tcolorbox}[
  breakable,
  enhanced,
  colback=gray!5!white,
  colframe=gray!60!black,
  title= User Profile Consistency,
]
\small{
\begin{lstlisting}[
  basicstyle=\ttfamily\small,
  breaklines=true,
  breakautoindent=false,
  breakindent=0pt,
  postbreak=
]
You are evaluating a user simulator in a conversation with an AI assistant.

You will be given:
- The user's profile (detailed ground truth user attributes)
- The full conversation history up to a specific turn

Your task is to judge whether the user's messages are consistent with their profile and attributes. A user is
consistent if the language, knowledge level, preferences, and behaviors they display in the
conversation match the persona described in the profile.

Output ONLY valid JSON. No markdown, no extra text. Use this format exactly:
{
    "result": <true or false>,
    "reason": "<brief reasoning>"
}

User profile:
{attributes}

Conversation history:
{conversation}

Return:

\end{lstlisting}
}
\end{tcolorbox}

\begin{tcolorbox}[
  breakable,
  enhanced,
  colback=gray!5!white,
  colframe=gray!60!black,
  title= Knowledge Consistency,
]
\small{
\begin{lstlisting}[
  basicstyle=\ttfamily\small,
  breaklines=true,
  breakautoindent=false,
  breakindent=0pt,
  postbreak=
]
You are evaluating a user simulator in a conversation with an AI assistant.

You will be given:
- The user's knowledge context (specific facts, background information the user is supposed to know)
- The full conversation history up to a specific turn

Your task is to judge whether the user's messages are consistent with their knowledge context.
A user is consistent if the facts and information they reference or rely on in their messages
match what is described in their knowledge context, without contradicting or inventing facts.

Output ONLY valid JSON. No markdown, no extra text. Use this format exactly:
{
    "result": <true or false>,
    "reason": "<brief reasoning>"
}

User knowledge context:
{knowledge}

Conversation history:
{conversation}

Return:

\end{lstlisting}
}
\end{tcolorbox}

% This is an appendix.

\end{document}